\documentclass{article}

\usepackage[preprint]{preprint}

\usepackage[utf8]{inputenc} 
\usepackage[T1]{fontenc}    
\usepackage{url}            
\usepackage{booktabs}       
\usepackage{amsfonts}       
\usepackage{nicefrac}       
\usepackage{microtype}      
\usepackage{graphicx}
\usepackage{amsmath}
\usepackage{makecell}
\usepackage{array}
\usepackage{tcolorbox}
\usepackage{enumitem}
\usepackage[table]{xcolor}
\usepackage{multirow}
\usepackage{colortbl}
\newcommand{\VarSty}[1]{\textcolor{blue}{#1}}

\definecolor{linkc}{rgb}{0, 0.44, 0.74}
\definecolor{eqc}{rgb}{1, 0, 0}
\definecolor{lightgray}{gray}{0.9}
\definecolor{datasetgray}{RGB}{242,244,247}
\definecolor{oursblue}{RGB}{226,239,249}

\newcommand{\datasetrow}[1]{%
\rowcolor{datasetgray}
\multicolumn{11}{c}{\textbf{\textsc{#1}}} \\
}

\usepackage[pagebackref=false,breaklinks=true,colorlinks=True,citecolor=linkc,linkcolor=eqc,bookmarks=false]{hyperref}

\usepackage{amsmath,amsfonts,bm}

\def\eqref#1{equation~\ref{#1}}

\def\1{\bm{1}}

\def\vc{{\bm{c}}}

\def\vp{{\bm{p}}}

\def\vt{{\bm{t}}}

\def\vz{{\bm{z}}}
\def\v\epsilon{{\bm{\epsilon}}}
\def\vI{{\bm{I}}}

\def\mR{{\bm{R}}}

\def\mT{{\bm{T}}}

\def\mW{{\bm{W}}}

\DeclareMathAlphabet{\mathsfit}{\encodingdefault}{\sfdefault}{m}{sl}
\SetMathAlphabet{\mathsfit}{bold}{\encodingdefault}{\sfdefault}{bx}{n}

\def\gS{{\mathcal{S}}}

\def\sR{{\mathbb{R}}}

\title{AnchorWorld: Embodied Egocentric World Simulation with View-based Evolution Customization}

\author{
  Yu Li\textsuperscript{1}\thanks{This work was conducted during the author’s internship at Kling Team, Kuaishou Technology.} \quad
  Menghan Xia\textsuperscript{2}\thanks{Corresponding authors.} \quad
  Gongye Liu\textsuperscript{4} \quad
  \textbf{Xintao Wang\textsuperscript{3}} \quad
  \textbf{Conglang Zhang\textsuperscript{5}} \\
  \textbf{Lei Ke\textsuperscript{1}} \quad
  \textbf{Yuxuan Lin\textsuperscript{1}} \quad
  \textbf{Ruihang Chu\textsuperscript{1}\footnotemark[2]} \quad
  \textbf{Pengfei Wan\textsuperscript{3}} \quad
  \textbf{Kun Gai\textsuperscript{3}} \quad
  \textbf{Yujiu Yang\textsuperscript{1}}
  \\[2ex]
  \textsuperscript{1}Tsinghua University \quad
  \textsuperscript{2}HUST \quad
  \textsuperscript{3}Kling Team, Kuaishou Technology \quad
  \textsuperscript{4}HKUST \quad
  \textsuperscript{5}WHU
  \\[1ex]
  \centerline{\url{https://yuli0103.github.io/AnchorWorld/}}
}

\begin{document}

\maketitle
\vspace{-0.7cm}

\begin{figure}[ht]
    \centering
    \includegraphics[width=\linewidth]{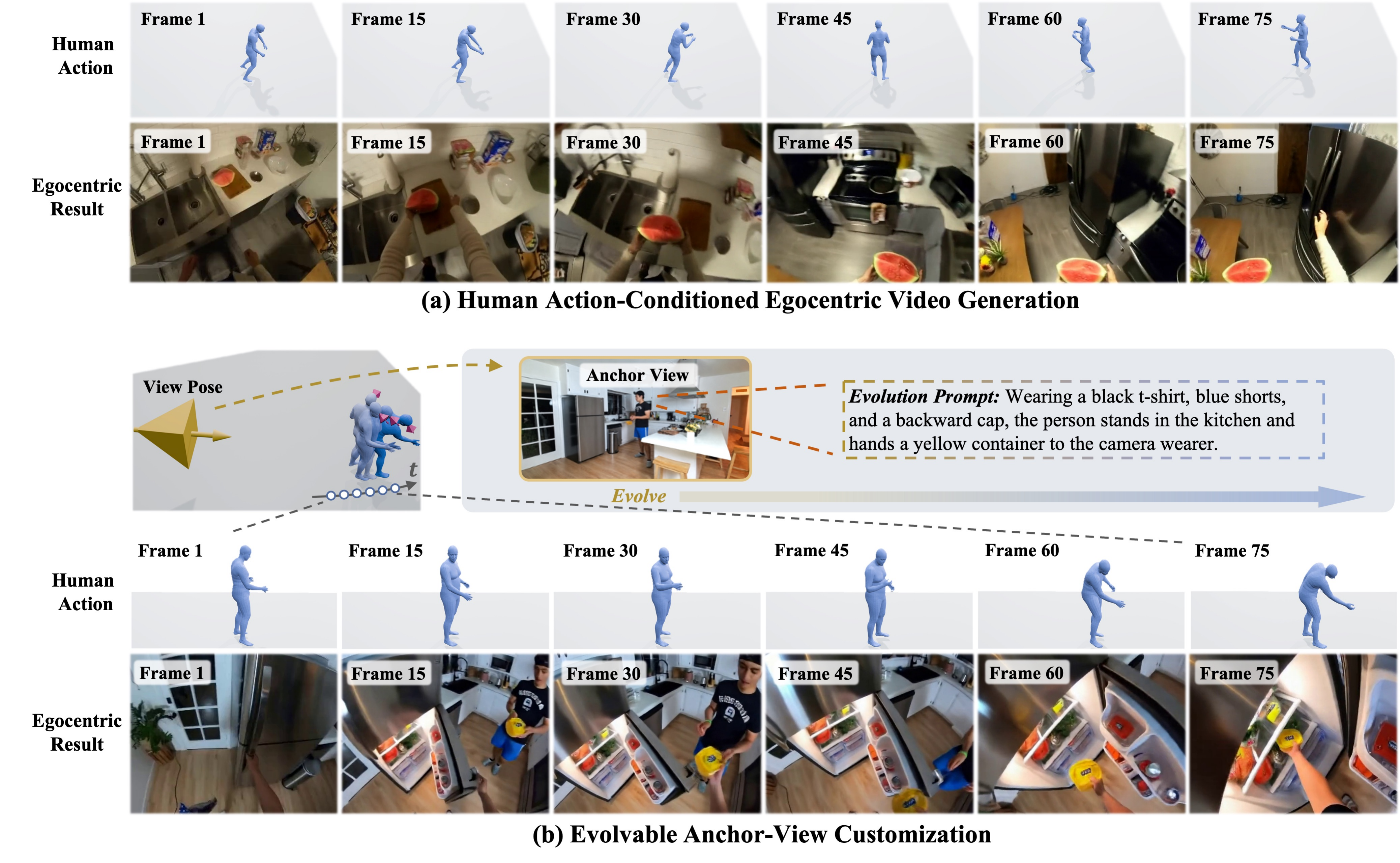}\vspace{-0.5em}
    \caption{
    Showcasing AnchorWorld. 
    (a) AnchorWorld synthesizes egocentric videos conditioned on human action and initial ego-view frame. 
    (b) It further enables world customization with conditional anchor views, which provide local appearance, 3D pose, and evolution prompts for scene evolution.
    }

    \label{fig:1}
\end{figure}

\begin{abstract}
Despite being a pivotal frontier, interactive world modeling remains underexplored in terms of the versatile controllability required by practical scenarios. To bridge this gap, we present AnchorWorld, a framework that advances egocentric simulation through enhanced interaction integrity and a flexible mechanism for world customization. First, we utilize 3D human motion as the primary interaction modality. To complement the out-of-view or truncated body parts in egocentric views, we introduce an auxiliary training supervision that incorporates exogenous viewpoints decoupled from the agent’s first-person sensorium. It allows the model to observe the agent's full-body positioning relative to the environment, facilitating a more robust spatial grounding of human-world interactions. Furthermore, we propose a simple yet effective mechanism for customizing self-evolving worlds. This is achieved by defining anchor views within a unified world coordinate system, coupled with textual descriptions dictating the dynamic evolution of local scenes. Experimental results show that AnchorWorld significantly outperforms state-of-the-art baselines, while ablation studies validate the effectiveness of our key designs. Notably, our customization scheme exhibits promising spatio-temporal geometric consistency and adheres strictly to the prescribed evolutionary dynamics.
\end{abstract}

\section{Introduction}\label{sec:intro}

Interactive world models aim to simulate dynamic visual environments that respond to user intervention. 
For first-person applications such as virtual reality and embodied AI~\cite{bar2025navigation, feng2025vidarc, gao2026dreamdojo}, this response is not merely a matter of predicting visually plausible continuations.
The simulator must account for how the user moves and acts: head motion determines where the camera looks, body motion drives navigation, and coordinated actions shape how the user interacts with nearby objects.
Meanwhile, the simulated world should not be treated as an unconstrained visual continuation: it should contain local states that can be specified, preserved, and evolved as the user moves through the environment.
Together, these requirements call for an egocentric world simulator with two complementary forms of control: \emph{embodied action control} and \emph{localized world-state customization}.

Existing interactive world models only partially satisfy these requirements. Many approaches~\cite{bruce2024genie, tang2025hunyuan, team2026advancing, yang2025longlive} rely on simplified control signals such as keyboard inputs, camera trajectories, or text prompts, which are convenient for navigation but do not reflect how humans act from a first-person perspective. 
Recent egocentric methods move toward more natural control by incorporating hand actions~\cite{wang2026hand2world, xie2026generated} or full-body motion~\cite{bai2025whole, tu2025playerone}.
However, learning such control from egocentric videos remains challenging.
The motion condition describes the body in 3D, while most of the body is absent from the egocentric frame to be predicted. 
Therefore, the model observes the visual consequences of body motion only indirectly, making motion supervision sparse and weakly aligned.
A second challenge lies in how the “world” itself is defined. 
Existing methods~\cite{tu2025playerone,yu2025context} typically determine the environment implicitly through an initial frame, a global prompt, or historical context; newly observed regions are therefore weakly constrained. 
This makes it difficult to specify what should exist at particular 3D locations or how local scene states should evolve over time.

The two limitations above motivate \textbf{AnchorWorld}, a framework for world-customizable embodied egocentric simulation. 
AnchorWorld provides two complementary forms of control: human body motion for egocentric navigation and interaction, and pose-associated anchor views for explicit world customization. 
For egocentric action control, the supervision missing from first-person videos is precisely what third-person videos provide, since the body and its interaction with the scene are visible from outside.
We thus pair 3D human motion with camera viewpoint and formulate action conditioning in a projection-based manner, where the camera viewpoint can correspond to either an external observation view or the head-mounted view, enabling hybrid-view training.
This \textit{hybrid-view human action control} lets the model learn how full-body motion shapes first-person visual observations.
%
For world customization, we represent local world states with \textit{pose-associated anchor views}. Each anchor view consists of an RGB image specifying local visual appearance, a 3D pose that grounds the anchor, and an evolution prompt that describes its dynamic changes. 
These anchors allow users to specify local states at chosen 3D locations, preserve them across changing viewpoints, and guide their evolution, including in regions initially out of sight.


%
We train AnchorWorld with a progressive strategy that introduces hybrid-view human action control, anchor-view scene consistency, and dynamic evolution in successive stages so each component builds on a stable base.
Across egocentric, synthetic UE, and captured real-world scenarios, AnchorWorld improves over adapted baselines on action accuracy, scene consistency, and dynamic evolution.
The results further reveal remarkable generalization to out-of-distribution scenarios, especially under large viewpoint changes and limited overlap between the initial ego-view and anchor views.
Additional analyses show two key capabilities for localized world customization: out-of-sight scene evolution and pose-consistent anchoring under spatial transformations.
Our contributions are summarized:

\begin{itemize}[leftmargin=2em]
    \item We formulate \emph{world-customizable embodied egocentric simulation}, a task that enables human-motion-driven exploration and interaction within customizable, self-evolving worlds.

    \item We propose \emph{AnchorWorld}, a unified framework that combines embodied egocentric action control with pose-associated anchor-view customization.
    

    \item We validate AnchorWorld through extensive experiments, demonstrating accurate egocentric human action control, strong spatial awareness, and controllable scene evolution.
\end{itemize}

\section{Related Work}\label{sec:2}


\paragraph{Interactive World Models.} 
The core pursuit of interactive world models is to synthesize visual environments conditioned on user input actions.
A large body of early research adopts keyboard and mouse operations to control viewpoints and navigate simulated worlds~\cite{bruce2024genie, hong2025relic, sun2025worldplay, team2026advancing, wang2026matrix, ye2025yan, zhu2025astra}.
Concurrently, another line of work employs text prompts as interaction signals, enabling users to trigger specific world events and drive environmental transitions~\cite{agarwal2025cosmos, chi2025wow, mao2025yume, shen2026egoforge, tang2025hunyuan, xiang2025pan, yang2025longlive}.
To support more fine-grained and embodied interactions, recent studies introduce hand poses as control signals~\cite{gao2026lome, hao2026egosim, li2026egocentric, wang2026hand2world, xie2026generated, zhang2026controllable}.
However, they are often limited to egocentric scenarios with restricted camera motion.
DWM~\cite{kim2025dexterous} performs interaction within static 3D scenes and achieves embodied simulation conditioned on rendered first-person videos and rendered hand meshes.
PlayerOne~\cite{tu2025playerone} uses full-body human motion to build egocentric world simulators.
It introduces a part-disentangled motion injection scheme, allowing the model to perceive the roles of different body parts.
Similarly, PEVA~\cite{bai2025whole} adopts human motion as the action condition and generates videos without text input, encouraging intention inference from first-person videos and motion cues.


\paragraph{Scene-Consistent Video Generation.} 

ReCamMaster~\cite{bai2025recammaster} tackles novel camera trajectory synthesis by enforcing scene consistency through source-video conditioning via in-context learning.
It further constructs paired training data with different camera trajectories using synthetic Unreal Engine data.
CineScene~\cite{huang2026cinescene} represents a scene with a dense sequence of images captured at regular angular intervals, and leverages implicit 3D features~\cite{wang2025vggt} to build scene understanding for camera-controlled cinematic video generation.
SWM~\cite{seo2026grounding} grounds its world model in real-world urban environments by retrieving nearby street-view images during navigation, and uses geometric and semantic references to improve spatial realism.
Context-as-Memory~\cite{yu2025context} maintains scene consistency in long video navigation by retrieving field-of-view-relevant historical frames and injecting both scene and viewpoint cues into generation.
Additionally, another line of work incorporates explicit 3D representations to improve view consistency across generated frames~\cite{fridman2023scenescape, huang2026gen3r, huang2025voyager, ni2025recondreamer, ren2025gen3c, yu2025wonderworld, yu2025trajectorycrafter, yu2024viewcrafter}.
These methods typically reconstruct or maintain intermediate 3D scene representations, such as depth maps or point clouds, and use them to guide novel-view or trajectory-conditioned video generation.



\section{Method}\label{sec:3}

\begin{figure}[!t]
    \centering
    \includegraphics[width=\linewidth]{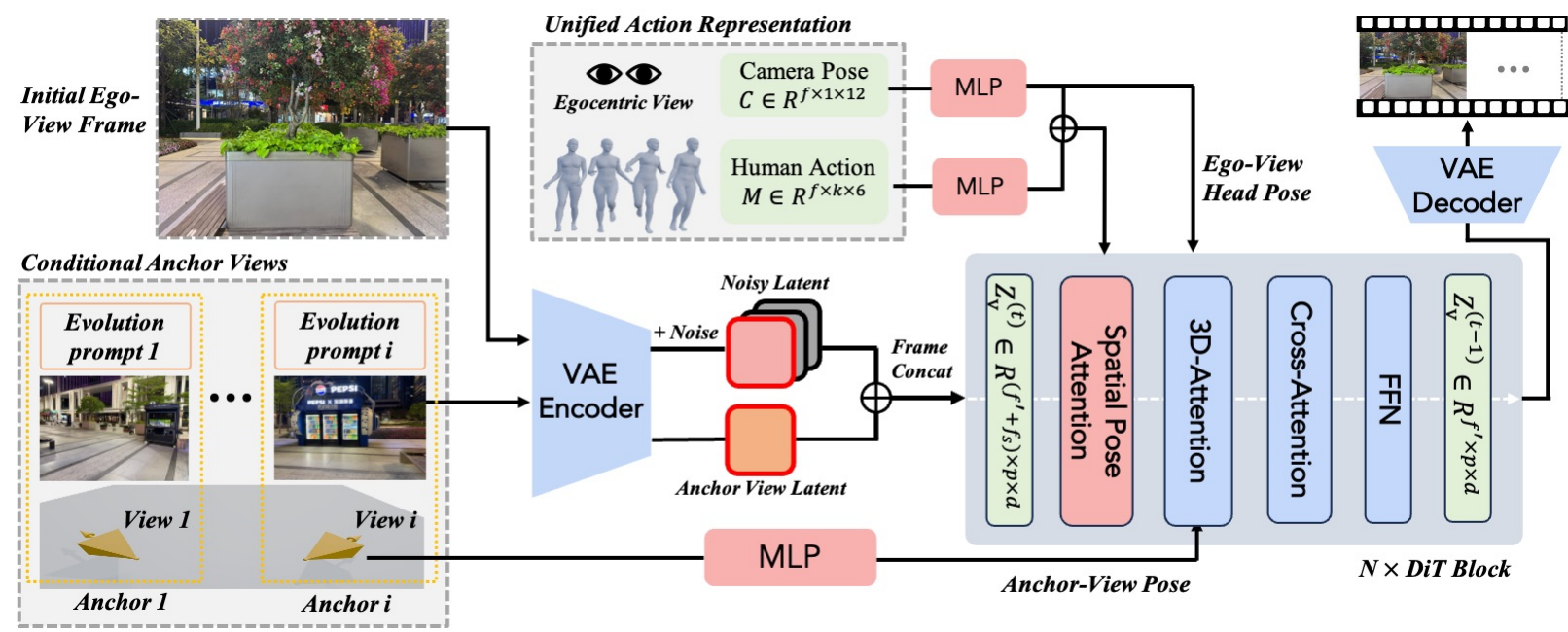}
    \caption{
    AnchorWorld synthesizes egocentric videos conditioned on embodied human actions and anchor views. 
    For action control, full-body motion and ego-view pose are concatenated as a unified action representation and injected via spatial pose attention. 
    For world customization, each anchor view includes an RGB image, a 3D pose, and an evolution prompt, enabling spatially grounded and temporally evolvable world simulation.
    Evolution prompts are incorporated via cross-attention layers.
    }
    \label{fig:method}
\end{figure}

Given a sequence of human actions and a customizable world specification, our goal is to synthesize an egocentric video that reflects how a user navigates and interacts within the defined environment. 
To this end, AnchorWorld takes two types of control signals as input: embodied human motion for action control, and pose-associated anchor views for world customization.
We instantiate AnchorWorld with Wan~\cite{wan2025wan}, a flow-matching-based~\cite{lipman2022flow} DiT~\citep{peebles2023scalable} video generation model, and condition its video synthesis on the action and anchor-view signals.
The human motion is represented as a sequence of body actions derived from the SMPL-X parametric body model~\cite{pavlakos2019expressive}, denoted as
$M \in \sR^{f \times k \times 6}$, where $f$ is the number of frames and $k$ is the number of joints. 
Each joint state consists of its 3D position and 3D axis-angle rotation vector. 
The customizable world is defined by an initial egocentric view $I_0$ and a set of localized anchor views $\gS = \{ (\vI_i, \vc_i, \vt_i) \}_{i=1}^n$. 
Each anchor view contains an RGB image $\vI_i$, a 6-DoF viewpoint pose $\vc_i = [\mR_i \mid \vp_i] \in \sR^{3 \times 4}$, and an evolution prompt $\vt_i$ that describes the temporal change of local scene states.
Figure~\ref{fig:method} provides an overview of the proposed framework. 
We detail each component of our approach in the following subsections.


\subsection{Hybrid-View Human Action Control}\label{sec:3.1}


\paragraph{Enhanced Egocentric Action Control via hybrid views.}
Human action contains rich spatial and interaction cues: the root trajectory determines global navigation, the limbs indicate potential interactions with the surrounding scene, and the head motion induces the egocentric viewpoint.
However, in first-person videos, most body parts are often outside the camera field of view, making direct supervision of full-body action control sparse and incomplete.
To overcome this limitation, we introduce third-person view (TPV) videos as auxiliary training data, where the full human body and its interactions with the surrounding scene are explicitly visible.
These videos provide rich interaction context and complete motion supervision, helping the model learn stronger spatial grounding between human motion and scene responses.
To support joint training on both TPV and first-person view (FPV) data within a unified framework, we formulate action conditioning in a projection-based manner.
Specifically, we represent the action condition by combining the full-body motion sequence with the camera trajectory, allowing the model to project 3D human motion into 2D visual observations under arbitrary viewpoints.
We first pre-train the model on large-scale and diverse TPV videos, where the camera parameters correspond to the external observation viewpoint, enabling the model to acquire projection knowledge and human-scene interaction priors.
Then, we adapt the model to egocentric simulation by aligning the camera parameters with the human head perspective in FPV data.
This design enables more accurate human-action control and stronger spatial pose awareness.

\paragraph{Spatial Pose Attention.} 
Inspired by prior work~\cite{fu20243dtrajmaster, li2025adaviewplanner}, we inject the pose conditions through a spatial pose attention mechanism. 
Specifically, a motion encoder first projects the input motion sequence $M \in \sR^{f \times k \times 6}$ into a latent embedding $\vz_m \in \sR^{f' \times k \times d}$, where $d$ is the model's hidden dimension. To ensure temporal alignment with the VAE-encoded~\cite{kingma2013auto} video latents, we employ temporal downsampling to match the temporal resolution $f'$. Analogously, a camera encoder processes the camera pose sequence $C \in \sR^{f \times 3 \times 4}$ into $\vz_c \in \sR^{f' \times 1 \times d}$, where the camera pose can represent either a third-person observation viewpoint or the first-person head viewpoint.

To exploit the inherent frame-wise correspondence between motion and video tokens, we concatenate the video tokens $\vz_v^{(t)}$ with the human motion tokens $\vz_m$ and camera pose tokens $\vz_c$ along the spatial dimension. This unified sequence is then processed by the spatial self-attention block:
% ...

\[
\begin{gathered}
\mT = [\vz_v^{(t)}; \vz_m; \vz_c] \in \sR^{f' \times (h \cdot w + k + 1) \times d}, \\
\vz_v^{(t)} = \vz_v^{(t)} + \text{Truncate}\left(\text{Attn}(\mW_Q \cdot \mT, \mW_K \cdot \mT, \mW_V \cdot \mT)\right)
\end{gathered}
\tag{1}
\]

The Truncate operator discards the auxiliary pose tokens, retaining only the updated video features.

\subsection{Evolvable Anchor-View Customization}\label{sec:3.2}

To enable evolvable world customization, we represent the environment with a set of anchor views. 
Each anchor view provides three types of localized world priors: an RGB image for visual appearance, a 3D pose for spatial grounding, and an evolution prompt for temporal state evolution. 

\paragraph{In-Context Anchor-View Priors.}
To incorporate anchor-view image priors while preserving the generative capability of the pre-trained video model, we adopt an in-context conditioning strategy~\cite{huang2026cinescene, ju2025fulldit, ye2025unic}. 
Specifically, the images of anchor views are encoded into latent tokens $\vz_s \in \sR^{f_s \times h \cdot w \times d}$, which are concatenated with the video latent tokens $\vz_v^{(t)} \in \sR^{f' \times h \cdot w \times d}$ along the frame dimension:
\[
    \mathcal{T}_{total} = [\vz_v^{(t)}; \vz_s] \in \sR^{(f' + f_s) \times h \cdot w \times d}. \tag{2}
\]

This design enables anchor views to guide world synthesis in-context, without requiring architectural modifications to the base model. We further employ 3D RoPE~\cite{su2024roformer} to differentiate anchor views by assigning them distinct frame-axis positions in the positional embedding space.

\begin{figure}[!t]
    \centering
    \includegraphics[width=\linewidth]{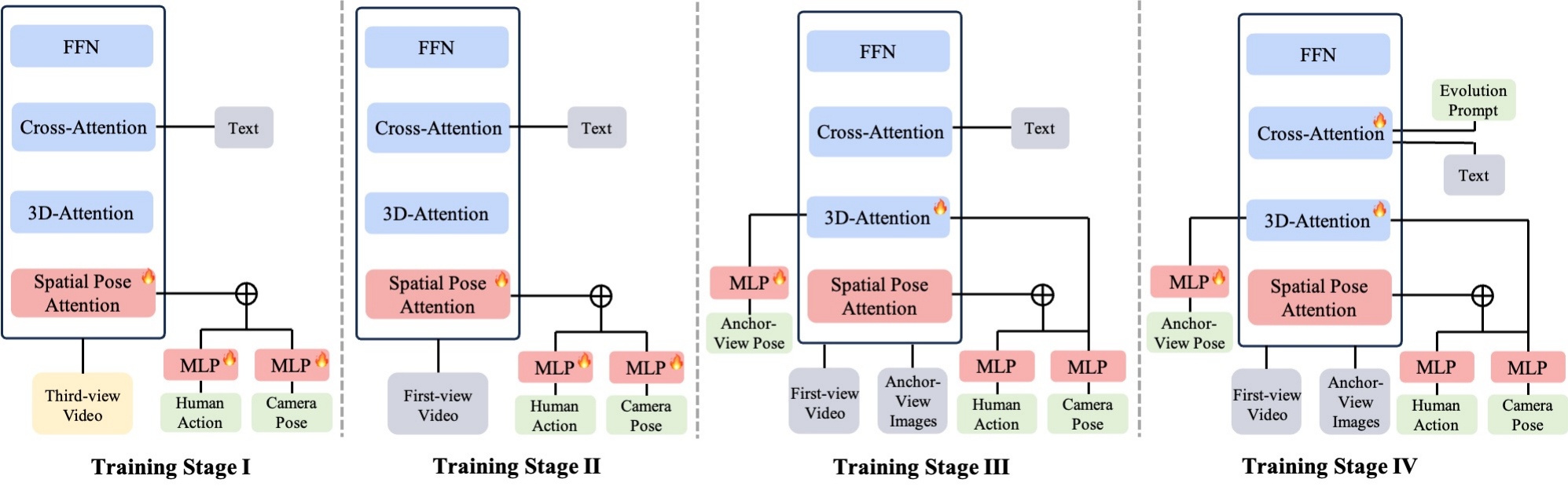}
    \caption{
        Progressive multi-stage training strategy.
        Stage I: TPV action training;
        Stage II: FPV action training;
        Stage III: static anchor-view customization;
        Stage IV: dynamic anchor-view evolution.
        }

    \label{fig:train}
\end{figure}

\paragraph{View Pose Injection.}
Since each view corresponds to a specific 3D location in the world, its spatial pose is essential for grounding the customized content. 
We therefore inject pose information for both generated video frames and anchor views. 
The camera poses are encoded into embeddings $\vz_{pose} \in \sR^{(f' + f_s) \times 1 \times d}$ and spatially broadcast to match the latent resolution, yielding $\vz_{pose} \in \sR^{(f' + f_s) \times h \cdot w \times d}$. 
Before the self-attention layers, the pose embeddings are added to the visual tokens:
\[
    \mathcal{T}_{total} = \mathcal{T}_{total} + \vz_{pose}. \tag{3}
\]

By coupling visual tokens with spatial poses, the model can distinguish anchor views located at different positions and associate the generated egocentric trajectory with the correct local constraints. 

\paragraph{Text-Driven Anchor-View Evolution.}
To enable dynamic world customization, each anchor view is paired with a localized evolution description $\vt_i$ that specifies its temporal scene changes.
We inject these descriptions through cross-attention, leveraging the semantic priors of the pre-trained video model.
To preserve the locality of dynamic instructions, we restrict the interaction between text prompts and visual tokens using an attention mask. 
For a text prompt $\vt_j$, its text keys are visible only to the generated video tokens and the corresponding anchor-view tokens $\vz_s^{(j)}$:
\[
    \mathcal{M}(q, k_j) = 
    \begin{cases} 
    0, & \text{if } q \in \vz_v \text{ or } q \in \vz_s^{(j)}, \\
    -\infty, & \text{otherwise}.
    \end{cases} \tag{4}
\]
This masked cross-attention enables anchor-specific text control, allowing local scene states to evolve over time while reducing interference across different anchor views.



\subsection{Progressive Multi-Stage Training Strategy}\label{sec:3.3}

To progressively equip the model with egocentric human action control and evolvable anchor-view customization, we adopt a multi-stage training strategy, as illustrated in Figure~\ref{fig:train}. 
\textbf{Stage I \& II: Hybrid-View Action Control Training.}
We train the model to learn action-conditioned generation from hybrid viewpoints, where TPV videos provide complete full-body motion supervision.
In Stage I, the model is trained on large-scale third-person videos, where the camera parameters represent external observation viewpoints.
In Stage II, we then adapt the model to first-person videos by aligning the camera trajectory with the head pose of the character.
\textbf{Stage III \& IV: Evolvable Anchor-View Customization Training.}
After establishing action controllability, we train the model to incorporate anchor-view priors for world customization.
In Stage III, we train the model on static scenes to learn pose-aware anchor-view conditioning for consistent egocentric roaming.
In Stage IV, we mix in dynamic data with evolution descriptions to model text-driven local state changes.

\begin{figure}[!t]
    \centering
    \includegraphics[width=\linewidth]{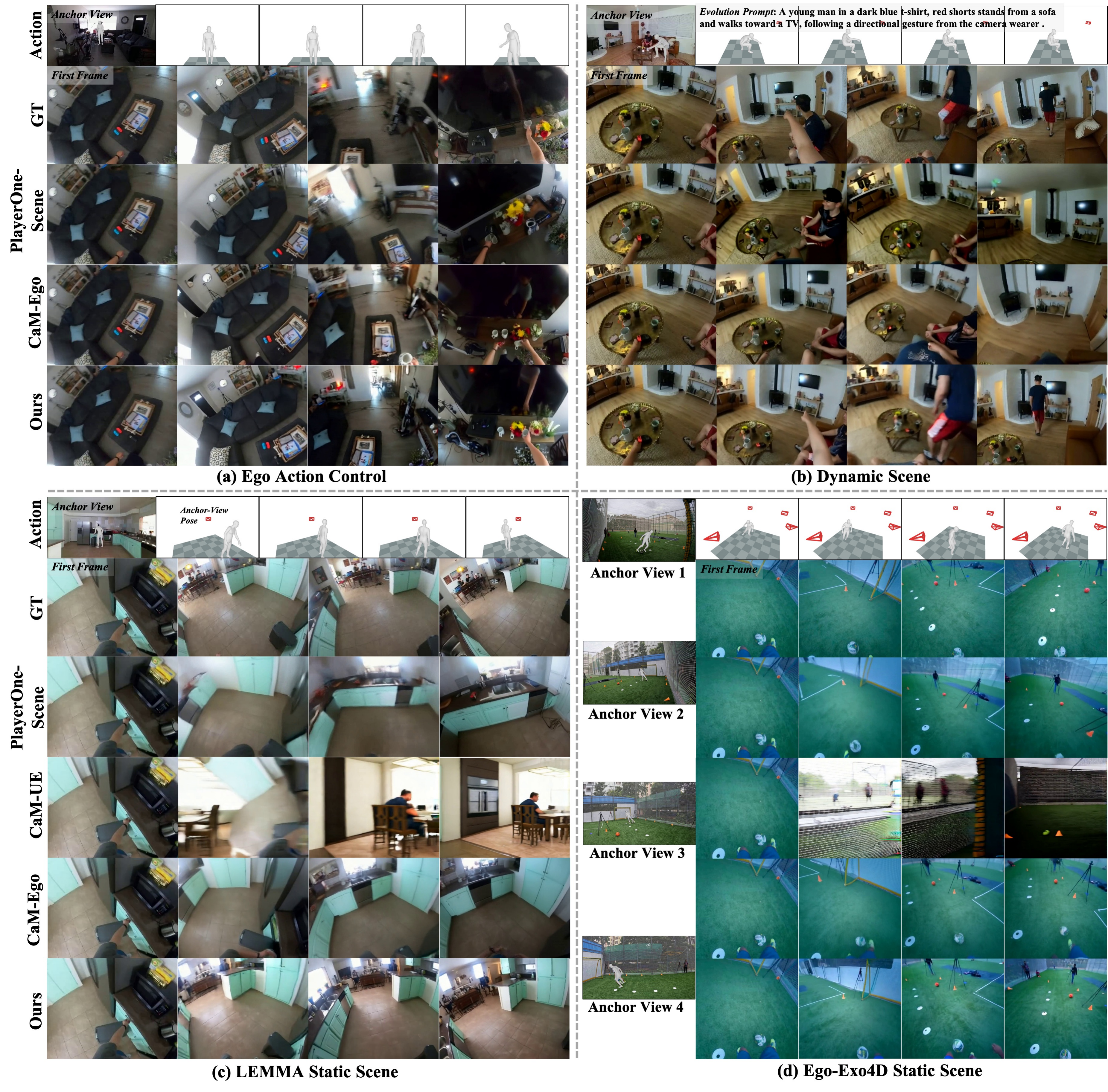}\vspace{-0.5em}
    \caption{Qualitative Comparison. The gray mask denotes the human action and its location in the anchor view. During inference, the gray-masked region in the anchor view is inpainted. Red wireframes visualize the 3D anchor-view poses. Our method achieves better egocentric action control, scene consistency under large viewpoint changes, and dynamic scene evolution.
}
    \label{fig:indomain_compare}
\end{figure}

\section{Experimental Results}\label{sec:4}

\subsection{Experiment Settings}\label{sec:4.1}

\noindent \textbf{Implementation Details.} 
We adopt Wan2.2 TI2V 5B~\cite{wan2025wan} as the base model and synthesize 77-frame videos at 480p resolution under an image-to-video formulation.
For exocentric training, we use an internally curated dataset of 200K single-person action videos and 101K videos from the UE-based MultiCamVideo dataset~\cite{bai2025recammaster}.
For egocentric training, we require synchronized third-person and first-person views to extract the camera wearer's human pose and anchor-view information; therefore, we use Ego-Exo4D~\cite{grauman2024ego} and LEMMA~\cite{jia2020lemma}, which provide synchronized cross-view pairings, diverse egocentric interactions, and dynamic scenes.
We use GVHMR~\cite{shen2024world} to estimate both 3D human motion and anchor-view poses in a shared 3D global coordinate system. The estimated motion contains 22 major body joints, excluding hand poses due to unreliable estimation in current egocentric data, as also noted in GigaHands~\cite{fu2025gigahands}. More details are provided in Appendix~\ref{app:a}.


\noindent \textbf{Baselines.}
We compare with baselines across three tasks:
(1) \textbf{Egocentric Action Control}: We use PlayerOne~\cite{tu2025playerone} as the main baseline, which decomposes human pose into body-part controls. Since its official code is unavailable, we re-implement it on Wan2.2 TI2V 5B using our training data, excluding hand poses for fairness due to unreliable estimation.
(2) \textbf{Static Scene Consistency}: We evaluate PlayerOne with our anchor-view injection mechanism, denoted as PlayerOne-Scene. We also compare with CaM~\cite{yu2025context}, which takes camera poses, scene context, and scene viewpoints as inputs, training two variants on our egocentric data and the official UE dataset. CineScene~\cite{huang2026cinescene} and SWM~\cite{seo2026grounding} are excluded due to FOV issues and unavailable code, respectively.
(3) \textbf{Dynamic Scene Evolution}: Since no prior work shares the same setting, we adapt the static-scene baselines by appending evolution prompts to their global text prompts.


\noindent \textbf{Evaluation.}
We evaluate the generated results from four aspects:
(1) \textbf{Action Accuracy}: As most body parts are out of view in egocentric videos, we quantify action controllability through camera-view control and qualitatively assess limb motion. 
Following MegaSaM~\cite{li2025megasam}, we use camera pose error metrics, including Absolute Translation Error (ATE), Relative Translation Error (RTE), and Relative Rotation Error (RRE). We estimate camera trajectories from synthesized videos using MegaSaM.
(2) \textbf{Scene Consistency}: Following prior works~\cite{bai2025recammaster, huang2026cinescene}, we report GIM-based Mat. Pix.~\cite{shen2024gim} to measure the ratio of matched pixels, CLIP-V~\cite{radford2021learning} for semantic similarity, pixel-aligned metrics including PSNR and SSIM~\cite{wang2004image}, and the perceptual metric LPIPS~\cite{zhang2018unreasonable}.
(3) \textbf{Dynamic Evolution}: We adopt the Text Alignment (TA) metric from VideoAlign~\cite{liu2025improving} to measure semantic consistency with anchor-view evolution prompts.
(4) \textbf{Video Quality}: We evaluate visual quality using VBench~\cite{huang2024vbench}. Averaged results are reported in Table~\ref{tab:quantitative_comparison}, with detailed results for each evaluation dimension provided in Table~\ref{tab:vbench_quantitative_comparison}.


\noindent \textbf{Test Sets.}
To evaluate performance and generalization, we construct four test sets:
(1) \textbf{Egocentric Static Test Set}: 100 held-out sequences from the same data sources as the training set, featuring significant motion and viewpoint variations.
(2) \textbf{UE Test Set}: 100 Unreal Engine (UE)~\cite{UnrealEngine5} sequences filtered from CineScene~\cite{huang2026cinescene}, whose initial frames do not overlap with the provided anchor views. Since CineScene camera trajectories are repurposed as character head poses, we retain only in-place rotational motions to avoid unnatural poses while preserving large viewpoint changes. Thus, viewpoint accuracy is evaluated only by RRE, and scene consistency is assessed only using CLIP-V and GIM due to inconsistent camera intrinsics.
(3) \textbf{Real-World Test Set}: sequences captured from diverse real-world scenes with anchor views and human actions under large viewpoint changes, used only for qualitative evaluation due to unavailable ground-truth references.
(4) \textbf{Egocentric Dynamic Test Set}: 100 held-out sequences from training data with pronounced dynamic human activities. We do not include out-of-domain dynamic data, as collecting such data remains challenging.

\begin{table*}[!t]
\centering
\caption{
Quantitative results.
For static scenes, we evaluate scene consistency, camera accuracy, and video quality.
For dynamic scenes, we additionally report text alignment with the evolution prompts.
For the CineScene test set, due to its data characteristics, we report only the applicable metrics.
}
\label{tab:quantitative_comparison}
\renewcommand{\arraystretch}{1.1}
\resizebox{\textwidth}{!}{%
\begin{tabular}{l ccccc ccc c c}
\toprule
\textbf{Method} 
& \multicolumn{5}{c}{\textbf{Scene Consistency}} 
& \multicolumn{3}{c}{\textbf{Camera Accuracy}} 
& \multicolumn{1}{c}{\textbf{Text Alignment}} 
& \multicolumn{1}{c}{\textbf{Video Quality}} \\
\cmidrule(lr){2-6} \cmidrule(lr){7-9} \cmidrule(lr){10-10} \cmidrule(lr){11-11}
& Mat. Pix.(K)\(\uparrow\) & CLIP-V\(\uparrow\) & PSNR\(\uparrow\) & SSIM\(\uparrow\) & LPIPS\(\downarrow\)
& ATE\(\downarrow\) & RTE\(\downarrow\) & RRE\(\downarrow\)
& VideoAlign-TA\(\uparrow\)
& VBench\(\uparrow\) \\
\midrule

\datasetrow{Ego Static Scene}
PlayerOne             &3961.6  &0.845  &13.26  &0.459  &0.596  &0.131  &0.037  &3.741  & $-$  &0.734  \\
PlayerOne-Scene       &4334.8  &0.864  &14.38  &0.500  &0.545  &0.142  &0.032  &3.353  & $-$  &0.735  \\
CaM-UE                &3706.9  &0.804  &11.57  &0.448  &0.686  &0.163  &0.040  &3.590  & $-$  &0.729  \\
CaM-Ego               &4379.4  &0.872  &15.16  &0.554  &0.515  &0.125  &0.032  &3.207  & $-$  &\textbf{0.748}  \\
\rowcolor{oursblue} \textbf{Ours}         
                       &\textbf{4493.4}  &\textbf{0.885}  &\textbf{16.06}  &\textbf{0.578}  &\textbf{0.470}  &\textbf{0.112}  &\textbf{0.029}  &\textbf{3.145}  & $-$  &0.748  \\
\midrule

\datasetrow{UE Static CineScene}
PlayerOne             &3947.0  &0.787  & $-$ & $-$ & $-$ & $-$ & $-$ & 2.438 & $-$  &0.736  \\
PlayerOne-Scene       &4413.5  &0.802  & $-$ & $-$ & $-$ & $-$ & $-$ & 2.401 & $-$  &0.737  \\
CaM-UE                &4301.1  &\textbf{0.852}  & $-$ & $-$ & $-$ & $-$ & $-$ & 1.722 & $-$  &0.750  \\
CaM-Ego               &4429.1  &0.842  & $-$ & $-$ & $-$ & $-$ & $-$ & 2.009 & $-$  &\textbf{0.770}  \\
\rowcolor{oursblue} \textbf{Ours}         
                       &\textbf{4555.1}  &0.851  & $-$ & $-$ & $-$ & $-$ & $-$ &\textbf{1.656}  & $-$  &0.769  \\
\midrule

\datasetrow{Ego Dynamic Scene}
PlayerOne-Scene       &4455.4  &0.864  &14.24  &0.454  &0.583  &0.067  &0.017  &1.784  &0.449  &0.756  \\
CaM-UE                &4466.5  &0.856  &12.82  &0.462  &0.627  &0.083  &0.018  &\textbf{1.230}  &0.115  &0.770  \\
CaM-Ego               &4459.0  &0.871  &14.57  &0.501  &0.574  &0.083  &0.016  &1.636  &0.385  &0.770  \\
\rowcolor{oursblue} \textbf{Ours}         
                       &\textbf{4634.6}  &\textbf{0.899}  &\textbf{16.37}  &\textbf{0.555}  &\textbf{0.486}  &\textbf{0.048}  &\textbf{0.013}  &1.346  &\textbf{0.717}  &\textbf{0.774}  \\

\bottomrule
\end{tabular}%
}
\vspace{-1.0em}
\end{table*}
\begin{figure}[h]
    \centering
    \includegraphics[width=\linewidth]{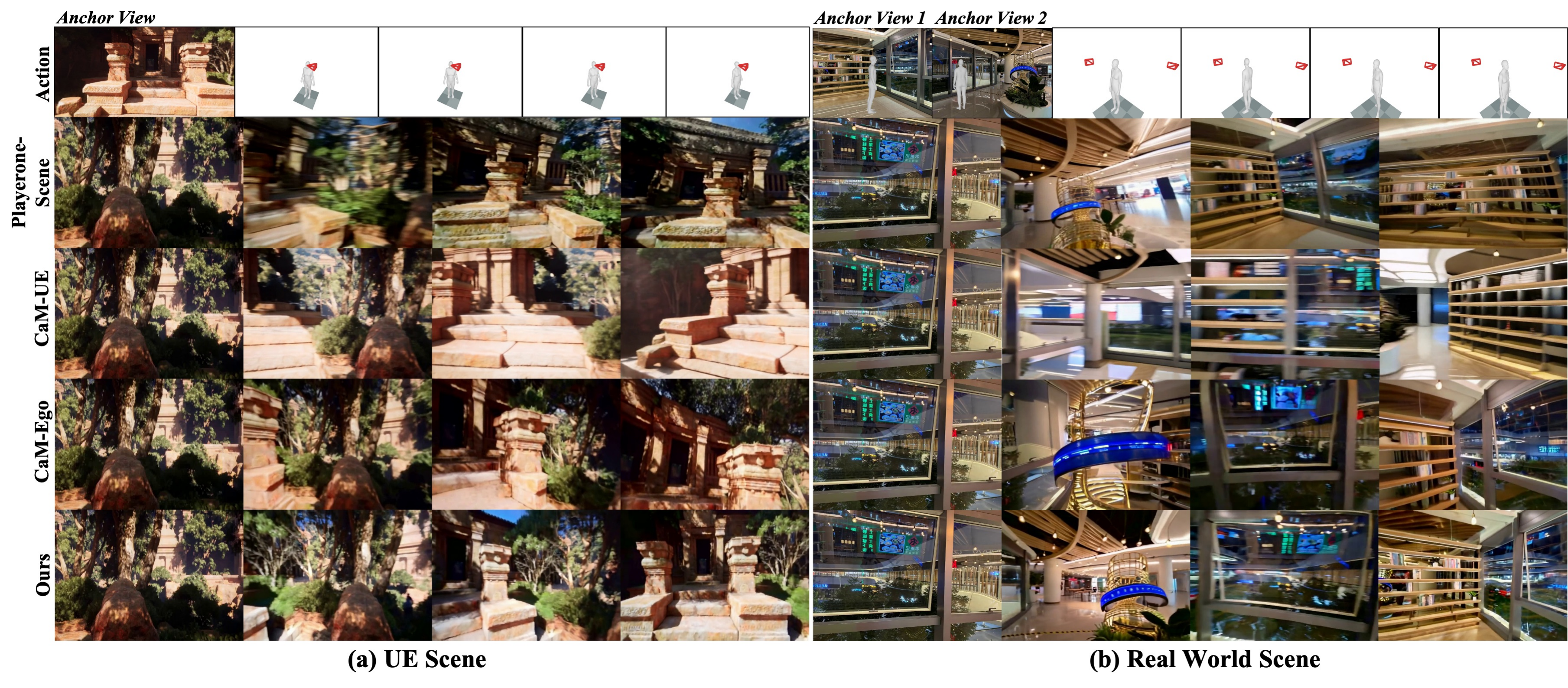}\vspace{-0.5em}
    \caption{Qualitative comparison on rendered UE scenes and real-world captured scenes.
}
    \label{fig:ood_compare}

\end{figure}

\subsection{Comparisons}\label{sec:4.2}

\noindent \textbf{Quantitative Results.}
As shown in Table~\ref{tab:quantitative_comparison}, our method achieves the best results in scene consistency, camera accuracy, and text alignment across all test scenarios, while maintaining comparable visual quality. 
PlayerOne learns from incomplete supervision targets captured from first-person videos, leading to weaker motion control. 
PlayerOne lacks scene information as input, and CaM-UE is trained only on UE data with slow camera motion; therefore, both methods perform poorly in scene consistency. 
Although PlayerOne-Scene and CaM-Ego are trained on the same data as ours, both exhibit weaker spatial perception than our projection-based control learning scheme. PlayerOne-Scene is limited by its part-wise learning scheme in modeling spatial pose variations, while CaM-Ego only takes viewpoint information as input.
Notably, since our method supports state evolution control via evolution prompts, its advantage becomes more pronounced in dynamic scenes.


\noindent \textbf{Qualitative Results.}
Figure~\ref{fig:indomain_compare} presents visual comparisons across multiple test tasks. 
Our method shows superior performance in egocentric human motion control, scene consistency under large viewpoint changes, and dynamic scene evolution driven by evolution prompts.
CaM-Ego only controls viewpoint changes without body motion input, while PlayerOne-Scene suffers from limited motion accuracy due to its part-wise control scheme. Additional action control results are shown in Figures~\ref{fig:indomain_action} and \ref{fig:real_world_action}. 
In addition, more results on evolution prompt control are shown in Figure~\ref{fig:dynamic_text_control}. 
We further evaluate our method on out-of-distribution UE scenes and real-world scenes in Figure~\ref{fig:ood_compare}, where there is limited or no overlap between the anchor view and the initial ego-view frame. The results show that our method exhibits strong generalization ability.
\subsection{Ablation Studies}\label{sec:4.3}

\noindent \textbf{Ablations of Design Strategies.}
We conduct ablation studies on key design choices in Table~\ref{tab:ablation_study}.
Under the action control setting, Stage-I third-person video training and the projection-based control design are essential, as shown quantitatively and visually in Figure~\ref{fig:indomain_action}.
Removing them also weakens scene consistency in later stages by degrading spatial perception.
In addition, removing anchor-view pose or anchor-view RoPE leads to worse scene consistency, confirming their roles in providing pose-aware view conditioning and distinguishing multiple anchor views.
Finally, we validate the effectiveness of the multi-stage training strategy through mixed-training variants across stages.


\begin{table*}[!t]
\centering
\caption{
Ablations on design choices.
We compare design strategies across egocentric action control and evolvable anchor-view customization.
TA denotes the VideoAlign text-alignment score.
``Joint'' denotes joint training of the corresponding stages.
}
\label{tab:ablation_study}
\renewcommand{\arraystretch}{1.1}
\resizebox{\textwidth}{!}{%
\begin{tabular}{l ccccc ccc c}
\toprule
\textbf{Variant} 
& \multicolumn{5}{c}{\textbf{Scene Consistency}} 
& \multicolumn{3}{c}{\textbf{Camera Accuracy}} 
& \multicolumn{1}{c}{\textbf{Text Alignment}} \\
\cmidrule(lr){2-6} \cmidrule(lr){7-9} \cmidrule(lr){10-10}
& Mat. Pix.(K)\(\uparrow\) & CLIP-V\(\uparrow\) & PSNR\(\uparrow\) & SSIM\(\uparrow\) & LPIPS\(\downarrow\)
& ATE\(\downarrow\) & RTE\(\downarrow\) & RRE\(\downarrow\)
& TA\(\uparrow\) \\
\midrule

\rowcolor{datasetgray}
\multicolumn{10}{c}{\small\textbf{Egocentric Action Control} \quad \textit{(Camera Accuracy)}} \\
w/o Stage I      & $-$ & $-$ & $-$ & $-$ & $-$  &0.125 &0.033 &3.532  & $-$ \\
w/o Head Pose     & $-$ & $-$ & $-$ & $-$ & $-$  &0.123 &0.032 &3.806  & $-$ \\
Joint Stage I \& II   & $-$ & $-$ & $-$ & $-$ & $-$  &0.123 &\textbf{0.030} &3.372  & $-$ \\
\rowcolor{oursblue} 
\textbf{Ours}                   & $-$ & $-$ & $-$ & $-$ & $-$  &\textbf{0.112} &0.030 &\textbf{3.187}  & $-$ \\
\midrule

\rowcolor{datasetgray}
\multicolumn{10}{c}{\small\textbf{Anchor-View Customization} \quad \textit{(Ego Static Scene)}} \\
w/o Stage I      &4438.3 &0.877 &15.50 &0.557 &0.492 &0.116 &0.031 &3.351 & $-$ \\
w/o Head Pose      &4425.4 &0.877 &15.42 &0.561 &0.502 &0.119 &0.032 &3.395 & $-$ \\
w/o Stage II      &4416.1 &0.879 &15.68 &0.571 &0.487 &0.115 &0.031 &3.234 & $-$ \\
w/o Anchor-View Pose            &4401.7 &0.879 &15.59 &0.568 &0.493 &0.112 &0.033 &3.184 & $-$ \\
w/o Anchor-View RoPE            &4395.2 &0.878 &15.40 &0.564 &0.498 &0.110 &0.031 &3.162 & $-$ \\
Joint Stage III \& IV   &4442.6 &0.877 &15.59 &0.570 &0.489 &\textbf{0.109} &0.031 &3.180 & $-$ \\
\rowcolor{oursblue} 
\textbf{Ours}                   &\textbf{4493.4} &\textbf{0.885} &\textbf{16.06} &\textbf{0.578} &\textbf{0.470} &0.112 &\textbf{0.029} &\textbf{3.145} & $-$ \\
\midrule

\rowcolor{datasetgray}
\multicolumn{10}{c}{\small\textbf{Anchor-View Evolution} \quad \textit{(Ego Dynamic Scene)}} \\
Joint Stage III \& IV   &4573.4 &0.893 &15.67 &0.522 &0.502 &0.050 &0.014 &1.362 &0.703 \\
\rowcolor{oursblue} 
\textbf{Ours}                   &\textbf{4634.6} &\textbf{0.899} &\textbf{16.37} &\textbf{0.555} &\textbf{0.486} &\textbf{0.048} &\textbf{0.013} &\textbf{1.346} &\textbf{0.717} \\

\bottomrule
\end{tabular}%
}
\vspace{-0.5em}
\end{table*}
\begin{figure}[!t]
    \centering
    \includegraphics[width=\linewidth]{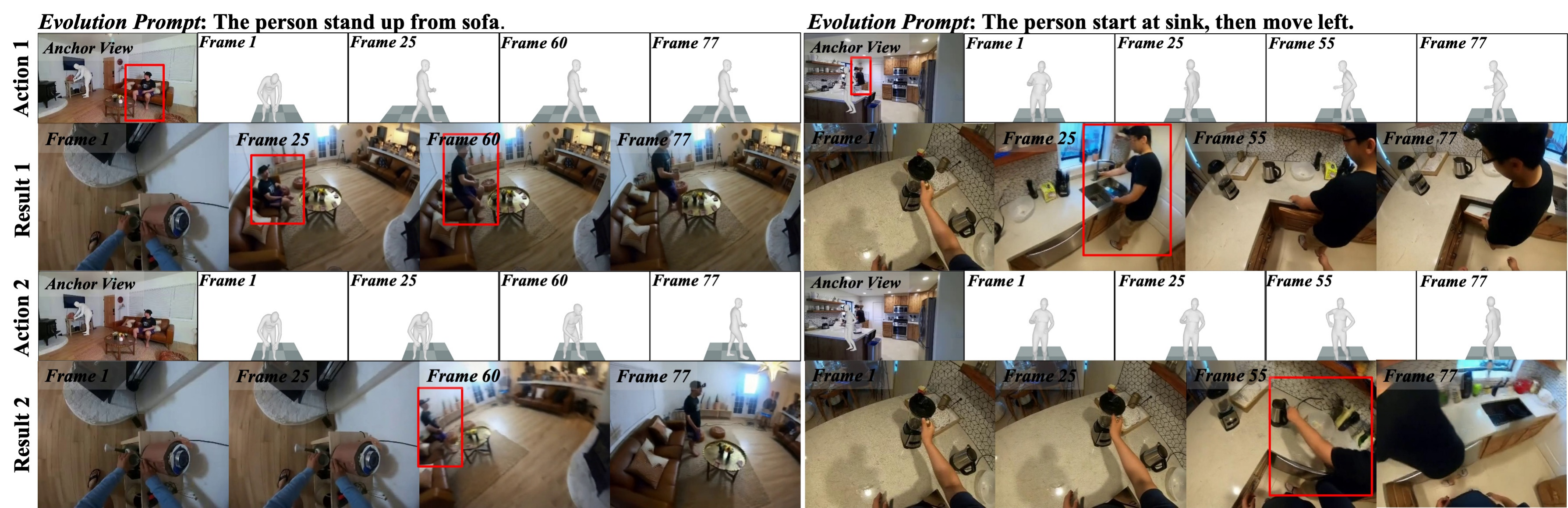}\vspace{-0.5em}
    \caption{Out-of-Sight Scene Evolution. We show that our model can infer scene evolution beyond the observed view by varying the timing of the action-induced viewpoint transition. Even when dynamic scene elements are not visible, our model can still reason about their state changes.
}
\vspace{-0.4cm}
    \label{fig:diff_pose}
\end{figure}

\noindent \textbf{Out-of-Sight Scene Evolution.}
As shown in Figure~\ref{fig:diff_pose}, we evaluate whether our model can infer scene dynamics beyond the initial egocentric view. We construct cases where another person appears in the anchor view but is initially invisible from the egocentric perspective, becoming visible only after a viewpoint change by the first-person player.
We vary the timing of this viewpoint change by modifying the egocentric human motion. For example, when the caption describes a person standing up from a sofa, an earlier viewpoint change reveals the person still sitting at frame~25 and subsequently standing up, whereas a delayed change first reveals them already standing at frame~60.


\begin{figure}[!t]
    \centering
    \includegraphics[width=\linewidth]{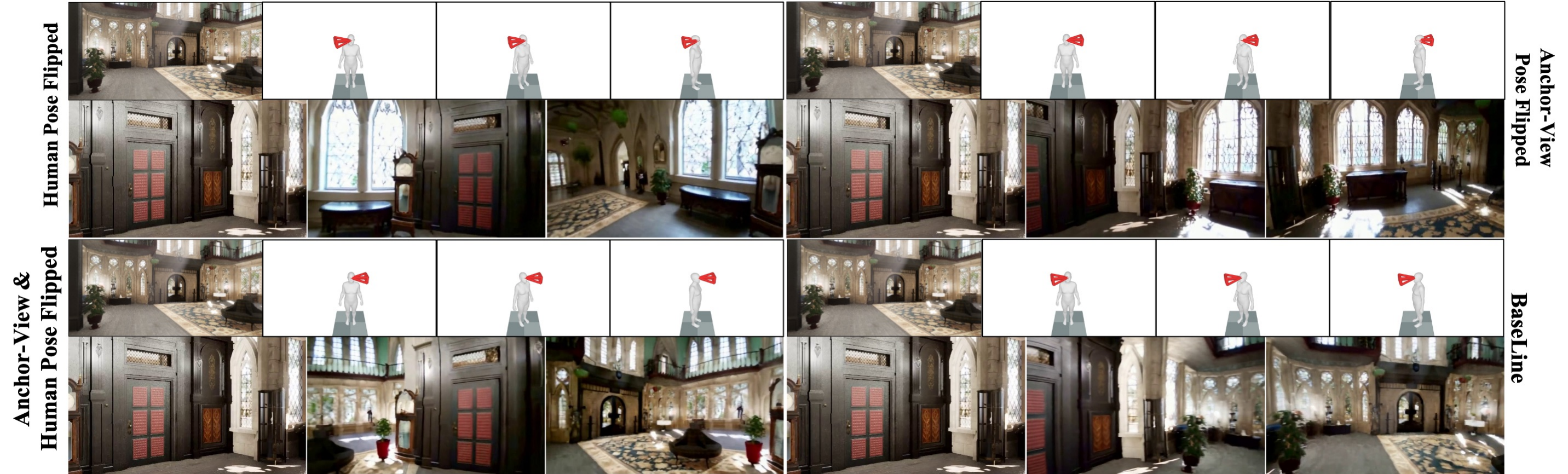}\vspace{-0.5em}
    \caption{Spatial Pose Awareness. We horizontally flip the human pose and anchor-view pose while keeping the anchor-view image fixed, creating overlapping and non-overlapping view settings.}
    \label{fig:spatial_understanding}
\end{figure}

\noindent \textbf{Spatial Pose Awareness.}
As shown in Figure~\ref{fig:spatial_understanding}, we flip the human and anchor-view poses, forming overlapping and non-overlapping settings. The results show that our method understands spatial pose relationships and retrieves appearance details when the poses overlap.


\section{Conclusion and Limitations}\label{sec:5}
In this work, we introduced AnchorWorld, a framework for world-customizable embodied egocentric simulation that integrates natural embodied action control with localized world-state customization.
Specifically, AnchorWorld leverages third-person videos to provide rich interaction context and complete human motion supervision, and employs projection-based action control to support hybrid-view training, while pose-associated anchor views provide spatially grounded appearance priors and text-driven local scene evolution.
Extensive experiments demonstrate that AnchorWorld consistently surpasses existing methods, while ablations validate each key design.
Additionally, AnchorWorld still has several limitations, including challenges in long-term exploration, open-world generalization, and diverse dynamic scenario modeling, which are discussed in detail in Appendix~\ref{app:b}.

\newpage

\bibliographystyle{plain}
\bibliography{references}

@inproceedings{bruce2024genie,
  title={Genie: Generative interactive environments},
  author={Bruce, Jake and Dennis, Michael D and Edwards, Ashley and Parker-Holder, Jack and Shi, Yuge and Hughes, Edward and Lai, Matthew and Mavalankar, Aditi and Steigerwald, Richie and Apps, Chris and others},
  booktitle={Forty-first International Conference on Machine Learning},
  year={2024}
}

@article{wang2026matrix,
  title={Matrix-Game 3.0: Real-Time and Streaming Interactive World Model with Long-Horizon Memory},
  author={Wang, Zile and Liu, Zexiang and Li, Jaixing and Huang, Kaichen and Xu, Baixin and Kang, Fei and An, Mengyin and Wang, Peiyu and Jiang, Biao and Wei, Yichen and others},
  journal={arXiv preprint arXiv:2604.08995},
  year={2026}
}

@article{xiang2025pan,
  title={Pan: A world model for general, interactable, and long-horizon world simulation},
  author={Xiang, Jiannan and Gu, Yi and Liu, Zihan and Feng, Zeyu and Gao, Qiyue and Hu, Yiyan and Huang, Benhao and Liu, Guangyi and Yang, Yichi and Zhou, Kun and others},
  journal={arXiv preprint arXiv:2511.09057},
  year={2025}
}

@article{hong2025relic,
  title={Relic: Interactive video world model with long-horizon memory},
  author={Hong, Yicong and Mei, Yiqun and Ge, Chongjian and Xu, Yiran and Zhou, Yang and Bi, Sai and Hold-Geoffroy, Yannick and Roberts, Mike and Fisher, Matthew and Shechtman, Eli and others},
  journal={arXiv preprint arXiv:2512.04040},
  year={2025}
}

@article{sun2025worldplay,
  title={Worldplay: Towards long-term geometric consistency for real-time interactive world modeling},
  author={Sun, Wenqiang and Zhang, Haiyu and Wang, Haoyuan and Wu, Junta and Wang, Zehan and Wang, Zhenwei and Wang, Yunhong and Zhang, Jun and Wang, Tengfei and Guo, Chunchao},
  journal={arXiv preprint arXiv:2512.14614},
  year={2025}
}

@article{zhu2025astra,
  title={Astra: General Interactive World Model with Autoregressive Denoising},
  author={Zhu, Yixuan and Feng, Jiaqi and Zheng, Wenzhao and Gao, Yuan and Tao, Xin and Wan, Pengfei and Zhou, Jie and Lu, Jiwen},
  journal={arXiv preprint arXiv:2512.08931},
  year={2025}
}

@article{ye2025yan,
  title={Yan: Foundational interactive video generation},
  author={Ye, Deheng and Zhou, Fangyun and Lv, Jiacheng and Ma, Jianqi and Zhang, Jun and Lv, Junyan and Li, Junyou and Deng, Minwen and Yang, Mingyu and Fu, Qiang and others},
  journal={arXiv preprint arXiv:2508.08601},
  year={2025}
}

@article{team2026advancing,
  title={Advancing Open-source World Models},
  author={Team, Robbyant and Gao, Zelin and Wang, Qiuyu and Zeng, Yanhong and Zhu, Jiapeng and Cheng, Ka Leong and Li, Yixuan and Wang, Hanlin and Xu, Yinghao and Ma, Shuailei and others},
  journal={arXiv preprint arXiv:2601.20540},
  year={2026}
}

@article{mao2025yume,
  title={Yume-1.5: A Text-Controlled Interactive World Generation Model},
  author={Mao, Xiaofeng and Li, Zhen and Li, Chuanhao and Xu, Xiaojie and Ying, Kaining and He, Tong and Pang, Jiangmiao and Qiao, Yu and Zhang, Kaipeng},
  journal={arXiv preprint arXiv:2512.22096},
  year={2025}
}

@article{tang2025hunyuan,
  title={Hunyuan-gamecraft-2: Instruction-following interactive game world model},
  author={Tang, Junshu and Liu, Jiacheng and Li, Jiaqi and Wu, Longhuang and Yang, Haoyu and Zhao, Penghao and Gong, Siruis and Yuan, Xiang and Shao, Shuai and Zhang, Linfeng and others},
  journal={arXiv preprint arXiv:2511.23429},
  year={2025}
}

@article{yang2025longlive,
  title={Longlive: Real-time interactive long video generation},
  author={Yang, Shuai and Huang, Wei and Chu, Ruihang and Xiao, Yicheng and Zhao, Yuyang and Wang, Xianbang and Li, Muyang and Xie, Enze and Chen, Yingcong and Lu, Yao and others},
  journal={arXiv preprint arXiv:2509.22622},
  year={2025}
}

@article{wang2026hand2world,
  title={Hand2world: Autoregressive egocentric interaction generation via free-space hand gestures},
  author={Wang, Yuxi and Ouyang, Wenqi and Wei, Tianyi and Dong, Yi and Shen, Zhiqi and Pan, Xingang},
  journal={arXiv preprint arXiv:2602.09600},
  year={2026}
}

@article{xie2026generated,
  title={Generated Reality: Human-centric World Simulation using Interactive Video Generation with Hand and Camera Control},
  author={Xie, Linxi and Sun, Lisong C and Neall, Ashley and Wu, Tong and Cai, Shengqu and Wetzstein, Gordon},
  journal={arXiv preprint arXiv:2602.18422},
  year={2026}
}

@article{hao2026egosim,
  title={EgoSim: Egocentric World Simulator for Embodied Interaction Generation},
  author={Hao, Jinkun and Jia, Mingda and Wang, Ruiyan and Liu, Xihui and Yi, Ran and Ma, Lizhuang and Pang, Jiangmiao and Xu, Xudong},
  journal={arXiv preprint arXiv:2604.01001},
  year={2026}
}

@article{li2026egocentric,
  title={Egocentric World Model for Photorealistic Hand-Object Interaction Synthesis},
  author={Li, Dayou and Liu, Lulin and Liu, Bangya and Zhou, Shijie and Feng, Jiu and Lu, Ziqi and Zheng, Minghui and You, Chenyu and Fan, Zhiwen},
  journal={arXiv preprint arXiv:2603.13615},
  year={2026}
}

@article{gao2026lome,
  title={LOME: Learning Human-Object Manipulation with Action-Conditioned Egocentric World Model},
  author={Gao, Quankai and Yang, Jiawei and Xu, Qiangeng and Chen, Le and Wang, Yue},
  journal={arXiv preprint arXiv:2603.27449},
  year={2026}
}

@article{tu2025playerone,
  title={Playerone: Egocentric world simulator},
  author={Tu, Yuanpeng and Luo, Hao and Chen, Xi and Bai, Xiang and Wang, Fan and Zhao, Hengshuang},
  journal={arXiv preprint arXiv:2506.09995},
  year={2025}
}

@inproceedings{yu2025context,
  title={Context as memory: Scene-consistent interactive long video generation with memory retrieval},
  author={Yu, Jiwen and Bai, Jianhong and Qin, Yiran and Liu, Quande and Wang, Xintao and Wan, Pengfei and Zhang, Di and Liu, Xihui},
  booktitle={Proceedings of the SIGGRAPH Asia 2025 Conference Papers},
  pages={1--11},
  year={2025}
}

@article{huang2026cinescene,
  title={CineScene: Implicit 3D as Effective Scene Representation for Cinematic Video Generation},
  author={Huang, Kaiyi and Huang, Yukun and Li, Yu and Bai, Jianhong and Wang, Xintao and Lin, Zinan and Ning, Xuefei and Yu, Jiwen and Wan, Pengfei and Wang, Yu and others},
  journal={arXiv preprint arXiv:2602.06959},
  year={2026}
}

@inproceedings{bai2025recammaster,
  title={Recammaster: Camera-controlled generative rendering from a single video},
  author={Bai, Jianhong and Xia, Menghan and Fu, Xiao and Wang, Xintao and Mu, Lianrui and Cao, Jinwen and Liu, Zuozhu and Hu, Haoji and Bai, Xiang and Wan, Pengfei and others},
  booktitle={Proceedings of the IEEE/CVF International Conference on Computer Vision},
  pages={14834--14844},
  year={2025}
}

@article{seo2026grounding,
  title={Grounding World Simulation Models in a Real-World Metropolis},
  author={Seo, Junyoung and Choi, Hyunwook and Kwon, Minkyung and Choi, Jinhyeok and Jin, Siyoon and Lee, Gayoung and Kim, Junho and Lee, JoungBin and Gu, Geonmo and Han, Dongyoon and others},
  journal={arXiv preprint arXiv:2603.15583},
  year={2026}
}

@inproceedings{pavlakos2019expressive,
  title={Expressive body capture: 3d hands, face, and body from a single image},
  author={Pavlakos, Georgios and Choutas, Vasileios and Ghorbani, Nima and Bolkart, Timo and Osman, Ahmed AA and Tzionas, Dimitrios and Black, Michael J},
  booktitle={Proceedings of the IEEE/CVF conference on computer vision and pattern recognition},
  pages={10975--10985},
  year={2019}
}

@article{li2025adaviewplanner,
  title={AdaViewPlanner: Adapting Video Diffusion Models for Viewpoint Planning in 4D Scenes},
  author={Li, Yu and Xia, Menghan and Liu, Gongye and Bai, Jianhong and Wang, Xintao and Zhang, Conglang and Lin, Yuxuan and Chu, Ruihang and Wan, Pengfei and Yang, Yujiu},
  journal={arXiv preprint arXiv:2510.10670},
  year={2025}
}

@article{ju2025fulldit,
  title={Fulldit: Multi-task video generative foundation model with full attention},
  author={Ju, Xuan and Ye, Weicai and Liu, Quande and Wang, Qiulin and Wang, Xintao and Wan, Pengfei and Zhang, Di and Gai, Kun and Xu, Qiang},
  journal={arXiv preprint arXiv:2503.19907},
  year={2025}
}

@article{ye2025unic,
  title={Unic: Unified in-context video editing},
  author={Ye, Zixuan and He, Xuanhua and Liu, Quande and Wang, Qiulin and Wang, Xintao and Wan, Pengfei and Zhang, Di and Gai, Kun and Chen, Qifeng and Luo, Wenhan},
  journal={arXiv preprint arXiv:2506.04216},
  year={2025}
}

@inproceedings{ren2025gen3c,
  title={Gen3c: 3d-informed world-consistent video generation with precise camera control},
  author={Ren, Xuanchi and Shen, Tianchang and Huang, Jiahui and Ling, Huan and Lu, Yifan and Nimier-David, Merlin and M{\"u}ller, Thomas and Keller, Alexander and Fidler, Sanja and Gao, Jun},
  booktitle={Proceedings of the IEEE/CVF Conference on Computer Vision and Pattern Recognition},
  pages={6121--6132},
  year={2025}
}

@article{fridman2023scenescape,
  title={Scenescape: Text-driven consistent scene generation},
  author={Fridman, Rafail and Abecasis, Amit and Kasten, Yoni and Dekel, Tali},
  journal={Advances in Neural Information Processing Systems},
  volume={36},
  pages={39897--39914},
  year={2023}
}

@article{yu2024viewcrafter,
  title={Viewcrafter: Taming video diffusion models for high-fidelity novel view synthesis},
  author={Yu, Wangbo and Xing, Jinbo and Yuan, Li and Hu, Wenbo and Li, Xiaoyu and Huang, Zhipeng and Gao, Xiangjun and Wong, Tien-Tsin and Shan, Ying and Tian, Yonghong},
  journal={arXiv preprint arXiv:2409.02048},
  year={2024}
}

@inproceedings{yu2025trajectorycrafter,
  title={Trajectorycrafter: Redirecting camera trajectory for monocular videos via diffusion models},
  author={Yu, Mark and Hu, Wenbo and Xing, Jinbo and Shan, Ying},
  booktitle={Proceedings of the IEEE/CVF international conference on computer vision},
  pages={100--111},
  year={2025}
}

@article{huang2025voyager,
  title={Voyager: Long-range and world-consistent video diffusion for explorable 3d scene generation},
  author={Huang, Tianyu and Zheng, Wangguandong and Wang, Tengfei and Liu, Yuhao and Wang, Zhenwei and Wu, Junta and Jiang, Jie and Li, Hui and Lau, Rynson and Zuo, Wangmeng and others},
  journal={ACM Transactions on Graphics (TOG)},
  volume={44},
  number={6},
  pages={1--15},
  year={2025},
  publisher={ACM New York, NY, USA}
}

@inproceedings{ni2025recondreamer,
  title={Recondreamer: Crafting world models for driving scene reconstruction via online restoration},
  author={Ni, Chaojun and Zhao, Guosheng and Wang, Xiaofeng and Zhu, Zheng and Qin, Wenkang and Huang, Guan and Liu, Chen and Chen, Yuyin and Wang, Yida and Zhang, Xueyang and others},
  booktitle={Proceedings of the Computer Vision and Pattern Recognition Conference},
  pages={1559--1569},
  year={2025}
}

@inproceedings{yu2025wonderworld,
  title={Wonderworld: Interactive 3d scene generation from a single image},
  author={Yu, Hong-Xing and Duan, Haoyi and Herrmann, Charles and Freeman, William T and Wu, Jiajun},
  booktitle={Proceedings of the Computer Vision and Pattern Recognition Conference},
  pages={5916--5926},
  year={2025}
}

@article{wan2025wan,
  title={Wan: Open and advanced large-scale video generative models},
  author={Wan, Team and Wang, Ang and Ai, Baole and Wen, Bin and Mao, Chaojie and Xie, Chen-Wei and Chen, Di and Yu, Feiwu and Zhao, Haiming and Yang, Jianxiao and others},
  journal={arXiv preprint arXiv:2503.20314},
  year={2025}
}

@inproceedings{grauman2024ego,
  title={Ego-exo4d: Understanding skilled human activity from first-and third-person perspectives},
  author={Grauman, Kristen and Westbury, Andrew and Torresani, Lorenzo and Kitani, Kris and Malik, Jitendra and Afouras, Triantafyllos and Ashutosh, Kumar and Baiyya, Vijay and Bansal, Siddhant and Boote, Bikram and others},
  booktitle={Proceedings of the IEEE/CVF Conference on Computer Vision and Pattern Recognition},
  pages={19383--19400},
  year={2024}
}

@inproceedings{jia2020lemma,
  title={Lemma: A multi-view dataset for le arning m ulti-agent m ulti-task a ctivities},
  author={Jia, Baoxiong and Chen, Yixin and Huang, Siyuan and Zhu, Yixin and Zhu, Song-Chun},
  booktitle={European Conference on Computer Vision},
  pages={767--786},
  year={2020},
  organization={Springer}
}

@inproceedings{shen2024world,
  title={World-grounded human motion recovery via gravity-view coordinates},
  author={Shen, Zehong and Pi, Huaijin and Xia, Yan and Cen, Zhi and Peng, Sida and Hu, Zechen and Bao, Hujun and Hu, Ruizhen and Zhou, Xiaowei},
  booktitle={SIGGRAPH Asia 2024 Conference Papers},
  pages={1--11},
  year={2024}
}

@article{bai2025qwen3,
  title={Qwen3-vl technical report},
  author={Bai, Shuai and Cai, Yuxuan and Chen, Ruizhe and Chen, Keqin and Chen, Xionghui and Cheng, Zesen and Deng, Lianghao and Ding, Wei and Gao, Chang and Ge, Chunjiang and others},
  journal={arXiv preprint arXiv:2511.21631},
  year={2025}
}

@inproceedings{fu2025gigahands,
  title={Gigahands: A massive annotated dataset of bimanual hand activities},
  author={Fu, Rao and Zhang, Dingxi and Jiang, Alex and Fu, Wanjia and Funk, Austin and Ritchie, Daniel and Sridhar, Srinath},
  booktitle={Proceedings of the Computer Vision and Pattern Recognition Conference},
  pages={17461--17474},
  year={2025}
}

@inproceedings{li2025megasam,
  title={Megasam: Accurate, fast and robust structure and motion from casual dynamic videos},
  author={Li, Zhengqi and Tucker, Richard and Cole, Forrester and Wang, Qianqian and Jin, Linyi and Ye, Vickie and Kanazawa, Angjoo and Holynski, Aleksander and Snavely, Noah},
  booktitle={Proceedings of the IEEE/CVF Conference on Computer Vision and Pattern Recognition},
  pages={10486--10496},
  year={2025}
}

@article{shen2024gim,
  title={Gim: Learning generalizable image matcher from internet videos},
  author={Shen, Xuelun and Cai, Zhipeng and Yin, Wei and M{\"u}ller, Matthias and Li, Zijun and Wang, Kaixuan and Chen, Xiaozhi and Wang, Cheng},
  journal={arXiv preprint arXiv:2402.11095},
  year={2024}
}

@inproceedings{radford2021learning,
  title={Learning transferable visual models from natural language supervision},
  author={Radford, Alec and Kim, Jong Wook and Hallacy, Chris and Ramesh, Aditya and Goh, Gabriel and Agarwal, Sandhini and Sastry, Girish and Askell, Amanda and Mishkin, Pamela and Clark, Jack and others},
  booktitle={International conference on machine learning},
  pages={8748--8763},
  year={2021},
  organization={PmLR}
}

@article{wang2004image,
  title={Image quality assessment: from error visibility to structural similarity},
  author={Wang, Zhou and Bovik, Alan C and Sheikh, Hamid R and Simoncelli, Eero P},
  journal={IEEE transactions on image processing},
  volume={13},
  number={4},
  pages={600--612},
  year={2004},
  publisher={IEEE}
}

@inproceedings{zhang2018unreasonable,
  title={The unreasonable effectiveness of deep features as a perceptual metric},
  author={Zhang, Richard and Isola, Phillip and Efros, Alexei A and Shechtman, Eli and Wang, Oliver},
  booktitle={Proceedings of the IEEE conference on computer vision and pattern recognition},
  pages={586--595},
  year={2018}
}

@article{liu2025improving,
  title={Improving video generation with human feedback},
  author={Liu, Jie and Liu, Gongye and Liang, Jiajun and Yuan, Ziyang and Liu, Xiaokun and Zheng, Mingwu and Wu, Xiele and Wang, Qiulin and Xia, Menghan and Wang, Xintao and others},
  journal={arXiv preprint arXiv:2501.13918},
  year={2025}
}

@inproceedings{huang2024vbench,
  title={Vbench: Comprehensive benchmark suite for video generative models},
  author={Huang, Ziqi and He, Yinan and Yu, Jiashuo and Zhang, Fan and Si, Chenyang and Jiang, Yuming and Zhang, Yuanhan and Wu, Tianxing and Jin, Qingyang and Chanpaisit, Nattapol and others},
  booktitle={Proceedings of the IEEE/CVF Conference on Computer Vision and Pattern Recognition},
  pages={21807--21818},
  year={2024}
}

@article{su2024roformer,
  title={Roformer: Enhanced transformer with rotary position embedding},
  author={Su, Jianlin and Ahmed, Murtadha and Lu, Yu and Pan, Shengfeng and Bo, Wen and Liu, Yunfeng},
  journal={Neurocomputing},
  volume={568},
  pages={127063},
  year={2024},
  publisher={Elsevier}
}

@article{kim2025dexterous,
  title={Dexterous World Models},
  author={Kim, Byungjun and Kim, Taeksoo and Lee, Junyoung and Joo, Hanbyul},
  journal={arXiv preprint arXiv:2512.17907},
  year={2025}
}

@inproceedings{wang2025vggt,
  title={Vggt: Visual geometry grounded transformer},
  author={Wang, Jianyuan and Chen, Minghao and Karaev, Nikita and Vedaldi, Andrea and Rupprecht, Christian and Novotny, David},
  booktitle={Proceedings of the Computer Vision and Pattern Recognition Conference},
  pages={5294--5306},
  year={2025}
}

@article{fu20243dtrajmaster,
  title={3dtrajmaster: Mastering 3d trajectory for multi-entity motion in video generation},
  author={Fu, Xiao and Liu, Xian and Wang, Xintao and Peng, Sida and Xia, Menghan and Shi, Xiaoyu and Yuan, Ziyang and Wan, Pengfei and Zhang, Di and Lin, Dahua},
  journal={arXiv preprint arXiv:2412.07759},
  year={2024}
}

@article{kingma2013auto,
  title={Auto-encoding variational bayes},
  author={Kingma, Diederik P and Welling, Max},
  journal={arXiv preprint arXiv:1312.6114},
  year={2013}
}

@inproceedings{peebles2023scalable,
  title={Scalable diffusion models with transformers},
  author={Peebles, William and Xie, Saining},
  booktitle={Proceedings of the IEEE/CVF international conference on computer vision},
  pages={4195--4205},
  year={2023}
}

@article{lipman2022flow,
  title={Flow matching for generative modeling},
  author={Lipman, Yaron and Chen, Ricky TQ and Ben-Hamu, Heli and Nickel, Maximilian and Le, Matt},
  journal={arXiv preprint arXiv:2210.02747},
  year={2022}
}

@misc{UnrealEngine5,
  author       = {{Epic Games}},
  title        = {Unreal Engine 5},
  howpublished = {\url{https://www.unrealengine.com/en-US/unreal-engine-5}},
  year         = {2022},
  note         = {Accessed: 2025-09-25}
}

@article{huang2026gen3r,
  title={Gen3R: 3D Scene Generation Meets Feed-Forward Reconstruction},
  author={Huang, Jiaxin and Yang, Yuanbo and Yang, Bangbang and Ma, Lin and Ma, Yuewen and Liao, Yiyi},
  journal={arXiv preprint arXiv:2601.04090},
  year={2026}
}

@article{zhang2025pretraining,
  title={Pretraining Frame Preservation in Autoregressive Video Memory Compression},
  author={Zhang, Lvmin and Cai, Shengqu and Li, Muyang and Zeng, Chong and Lu, Beijia and Rao, Anyi and Han, Song and Wetzstein, Gordon and Agrawala, Maneesh},
  journal={arXiv preprint arXiv:2512.23851},
  year={2025}
}

@article{bai2025whole,
  title={Whole-body conditioned egocentric video prediction},
  author={Bai, Yutong and Tran, Danny and Bar, Amir and LeCun, Yann and Darrell, Trevor and Malik, Jitendra},
  journal={arXiv preprint arXiv:2506.21552},
  year={2025}
}

@article{chi2025wow,
  title={Wow: Towards a world omniscient world model through embodied interaction},
  author={Chi, Xiaowei and Jia, Peidong and Fan, Chun-Kai and Ju, Xiaozhu and Mi, Weishi and Zhang, Kevin and Qin, Zhiyuan and Tian, Wanxin and Ge, Kuangzhi and Li, Hao and others},
  journal={arXiv preprint arXiv:2509.22642},
  year={2025}
}

@article{shen2026egoforge,
  title={EgoForge: Goal-Directed Egocentric World Simulator},
  author={Shen, Yifan and Liu, Jiateng and Li, Xinzhuo and Liu, Yuanzhe and Li, Bingxuan and Yang, Houze and Jia, Wenqi and Li, Yijiang and Yu, Tianjiao and Rehg, James Matthew and others},
  journal={arXiv preprint arXiv:2603.20169},
  year={2026}
}

@article{zhang2026controllable,
  title={Controllable Egocentric Video Generation via Occlusion-Aware Sparse 3D Hand Joints},
  author={Zhang, Chenyangguang and Ye, Botao and Chen, Boqi and Delitzas, Alexandros and Wang, Fangjinhua and Pollefeys, Marc and Wang, Xi},
  journal={arXiv preprint arXiv:2603.11755},
  year={2026}
}

@article{agarwal2025cosmos,
  title={Cosmos world foundation model platform for physical ai},
  author={Agarwal, Niket and Ali, Arslan and Bala, Maciej and Balaji, Yogesh and Barker, Erik and Cai, Tiffany and Chattopadhyay, Prithvijit and Chen, Yongxin and Cui, Yin and Ding, Yifan and others},
  journal={arXiv preprint arXiv:2501.03575},
  year={2025}
}

@inproceedings{bar2025navigation,
  title={Navigation world models},
  author={Bar, Amir and Zhou, Gaoyue and Tran, Danny and Darrell, Trevor and LeCun, Yann},
  booktitle={Proceedings of the Computer Vision and Pattern Recognition Conference},
  pages={15791--15801},
  year={2025}
}

@article{feng2025vidarc,
  title={Vidarc: Embodied Video Diffusion Model for Closed-loop Control},
  author={Feng, Yao and Xiang, Chendong and Mao, Xinyi and Tan, Hengkai and Zhang, Zuyue and Huang, Shuhe and Zheng, Kaiwen and Liu, Haitian and Su, Hang and Zhu, Jun},
  journal={arXiv preprint arXiv:2512.17661},
  year={2025}
}

@article{gao2026dreamdojo,
  title={DreamDojo: A Generalist Robot World Model from Large-Scale Human Videos},
  author={Gao, Shenyuan and Liang, William and Zheng, Kaiyuan and Malik, Ayaan and Ye, Seonghyeon and Yu, Sihyun and Tseng, Wei-Cheng and Dong, Yuzhu and Mo, Kaichun and Lin, Chen-Hsuan and others},
  journal={arXiv preprint arXiv:2602.06949},
  year={2026}
}

\newpage
\appendix
\section*{Appendix}
\appendix

The appendix consists of four sections. Readers can click on each section number to navigate to the corresponding section:
\begin{itemize}
    \item Section~\hyperref[app:a]{\textbf{A}} provides more implementation details.
    \item Section~\hyperref[app:b]{\textbf{B}} describes the limitations.
    \item Section~\hyperref[app:c]{\textbf{C}} describes additional analyses and results, including dynamic evolution prompt control, egocentric and exocentric action control, and real-world hard scenes.
    \item Section~\hyperref[app:d]{\textbf{D}} describes the failure cases.
\end{itemize} 

\section{Implementation Details}\label{app:a}

\begin{table}[!htbp]
\centering
\caption{
Overview of the progressive training stages.
}
\label{tab:training_stages}
\resizebox{\textwidth}{!}{%
\begin{tabular}{
l
>{\raggedright\arraybackslash}p{0.22\textwidth}
>{\raggedright\arraybackslash}p{0.20\textwidth}
>{\raggedright\arraybackslash}p{0.20\textwidth}
>{\raggedright\arraybackslash}p{0.22\textwidth}
}
\toprule
\textbf{Setting}
& \textbf{Stage I}
& \textbf{Stage II}
& \textbf{Stage III}
& \textbf{Stage IV} \\
\midrule

\textbf{Objective}
& Exocentric Motion
& Egocentric Motion 
& Static Scene
& Dynamic Scene \\
\addlinespace[2pt]

\textbf{Training Data}
& Internal videos; MultiCamVideo~\cite{bai2025recammaster}
& Ego-Exo4D~\cite{grauman2024ego}; LEMMA~\cite{jia2020lemma}
& Ego-Exo4D~\cite{grauman2024ego}; LEMMA~\cite{jia2020lemma}
& Ego-Exo4D~\cite{grauman2024ego}; LEMMA~\cite{jia2020lemma} \\
\addlinespace[2pt]

\textbf{Data Scale}
& 200K+101K
& 100K
& 25K
& 25K+10K \\
\addlinespace[2pt]

\textbf{Iterations}
& 30K
& 15K
& 10K
& 10K \\
\addlinespace[2pt]

\textbf{Batch Size}
& 16
& 16
& 16
& 16 \\
\addlinespace[2pt]

\textbf{Learning Rate}
& $1 \times 10^{-4}$
& $1 \times 10^{-4}$
& $1 \times 10^{-4}$
& $1 \times 10^{-4}$ \\
\addlinespace[2pt]

\textbf{Optimizer}
& AdamW
& AdamW
& AdamW
& AdamW \\
\addlinespace[2pt]

\textbf{Compute Resources}
& 16 NVIDIA GPUs@80G
& 16 NVIDIA GPUs@80G
& 16 NVIDIA GPUs@80G
& 16 NVIDIA GPUs@80G \\
\addlinespace[2pt]

\textbf{GPU Hours}
& 600 
& 300
& 253
& 253 \\

\bottomrule
\end{tabular}%
}
\vspace{2pt}
\end{table}

We adopt Wan2.2 TI2V 5B~\cite{wan2025wan} as our base video generation model and train it in an image-to-video manner.
Our training follows a progressive strategy, as summarized in Table~\ref{tab:training_stages}.
In the data scale row, Stage~I uses 200K internally curated real single-person action videos and 101K synthetic UE videos from MultiCamVideo~\cite{bai2025recammaster}; 
Stage II uses 100K egocentric action samples; Stage III uses 25K filtered samples with large viewpoint changes; and Stage IV jointly trains on the 25K static-scene samples from Stage III and 10K filtered dynamic-scene samples with noticeable human activities.

All videos are processed at 480p resolution while preserving their original aspect ratios, which retains visual content and avoids geometric distortion.
All training stages are conducted on 16 NVIDIA GPUs with a total batch size of 16, a learning rate of $1 \times 10^{-4}$, a timestep shift of 15, and the AdamW optimizer.
During training, pose conditions and anchor-view information are independently dropped with a probability of 5\%.
During inference, we use 50 denoising steps and set the classifier-free guidance scale to 5.

For egocentric video data, LEMMA~\cite{jia2020lemma} provides one anchor view for each sample, whereas Ego-Exo4D~\cite{grauman2024ego} contains one to six anchor views captured from different viewpoints. 
Due to the construction procedure of these paired third-person-to-first-person datasets, the anchor view images may contain the first-person player. Ideally, an anchor view should be defined independently of the player and thus should not include the player itself. 
However, given the relatively low data resolution, directly masking the player and applying inpainting would introduce visible artifacts and degrade image quality. Therefore, we do not apply inpainting during training. 
Importantly, using clean anchor-view images at inference time does not adversely affect the results. 
This can be attributed to two factors: (1) supervision from first-person videos enables the model to learn to ignore the player when interpreting anchor views; and (2) our input conditions include both human pose and view pose information, which allows the model to determine spatial relationships based primarily on pose cues.

For Ego-Exo4D, we undistort the egocentric fisheye videos and apply moderate brightness enhancement due to their low illumination. In addition, Ego-Exo4D exhibits noticeable color discrepancies between third-person and first-person videos, as these videos are captured by the different cameras. Nevertheless, our model can leverage valuable scene information from the anchor view while maintaining a color tone consistent with the initial ego-view frame.

For human motion and anchor-view pose estimation, we use GVHMR~\cite{shen2024world}. Specifically, we estimate 3D human motion from third-person views and canonicalize each sequence by placing the initial pose at the origin and aligning its horizontal orientation. The estimated motion contains 22 major body joints, excluding hand poses because hand estimation is unreliable in current egocentric data, due to frequent out-of-view hands, occlusions, and multi-person interference, as also noted in GigaHands~\cite{fu2025gigahands}. We further use GVHMR to estimate anchor-view poses relative to the target subject, thereby unifying human motion and anchor viewpoints in a shared 3D global coordinate system. In multi-person scenes, the estimated human motion may correspond to a non-egocentric subject. We therefore manually inspect the annotations, correct subject assignments when necessary, and discard samples with low-quality motion estimation.

For evolution prompts, they are annotated by Qwen3-VL-32B-Instruct~\cite{bai2025qwen3} using carefully designed prompt templates, as shown in Table~\ref{tab:fpv_character_description_template}.

\begin{figure}[!t]
    \centering
    \includegraphics[width=\linewidth]{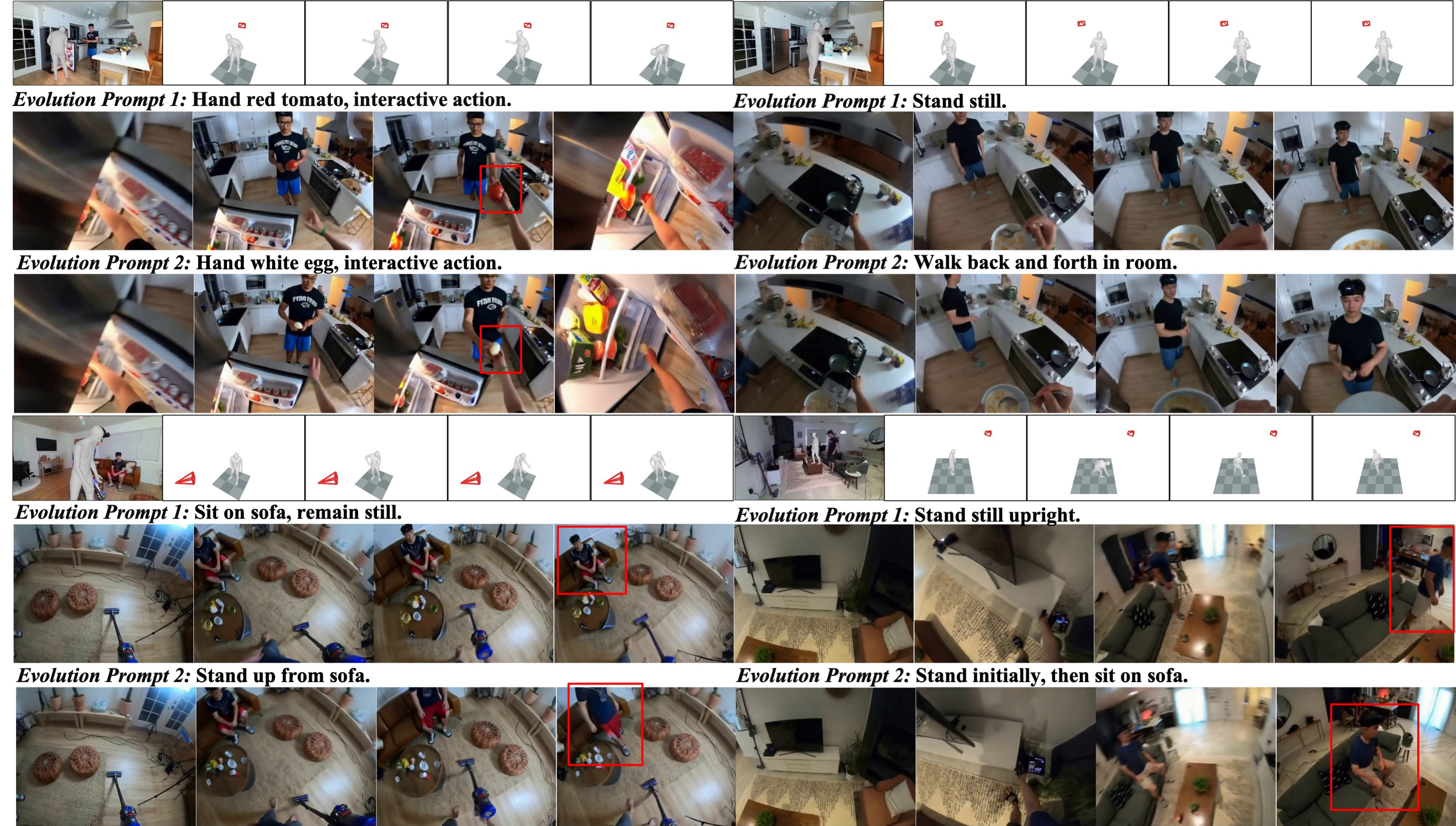}
    \caption{Evolution prompt control. We demonstrate that, within the same anchor-view image, modifying the evolution prompt enables control over different state changes.}
    \label{fig:dynamic_text_control}
\end{figure}

\section{Limitation}\label{app:b}

\noindent \textbf{Long-Term Exploration.}
In this work, we primarily focus on scenarios involving short video clips. However, enabling longer-horizon world exploration and interaction is essential for future progress. To this end, we plan to extend our framework toward real-time autoregressive interaction. We note that, in first-person settings, an embodied agent may continuously interact with the environment and explore larger-scale scenes. During this process, the model must update environmental state changes induced by its own actions in real time. Addressing this challenge requires a stronger emphasis on long-term memory mechanisms~\cite{zhang2025pretraining} within the model.

\noindent \textbf{Open World.} 
In this work, the training data primarily focuses on a constrained set of scenarios. In the future, collecting open-world data to construct broader environments and support longer-horizon world exploration will be an important direction.

\noindent \textbf{Diverse Dynamic Scenarios.}
Due to limitations in current egocentric training data, which typically provide multiple viewpoints of the same dynamic human activity, our empirical implementation uses a globally consistent evolution description for all anchor-view priors, i.e., $t_1 = \dots = t_n$, and mainly focuses on human-related activities rather than diverse dynamic scenarios. 
Future work can extend our framework to more diverse scenarios and anchor-specific dynamic controls, while incorporating the natural dynamic evolution of the world, thereby enabling the construction of more realistic and temporally rich worlds.

\section{Additional Analyses and Results}\label{app:c}

\subsection{Evolution Prompt Control}

As shown in Figure~\ref{fig:dynamic_text_control}, we achieve different dynamic evolutions of the scene by modifying the evolution prompt. This demonstrates that our method provides flexible support for diverse dynamic evolutions, allowing users to describe anchor-specific dynamic evolution.

\subsection{Egocentric Action Control}\label{app:d-2}

\begin{figure}[!t]
    \centering
    \includegraphics[width=\linewidth]{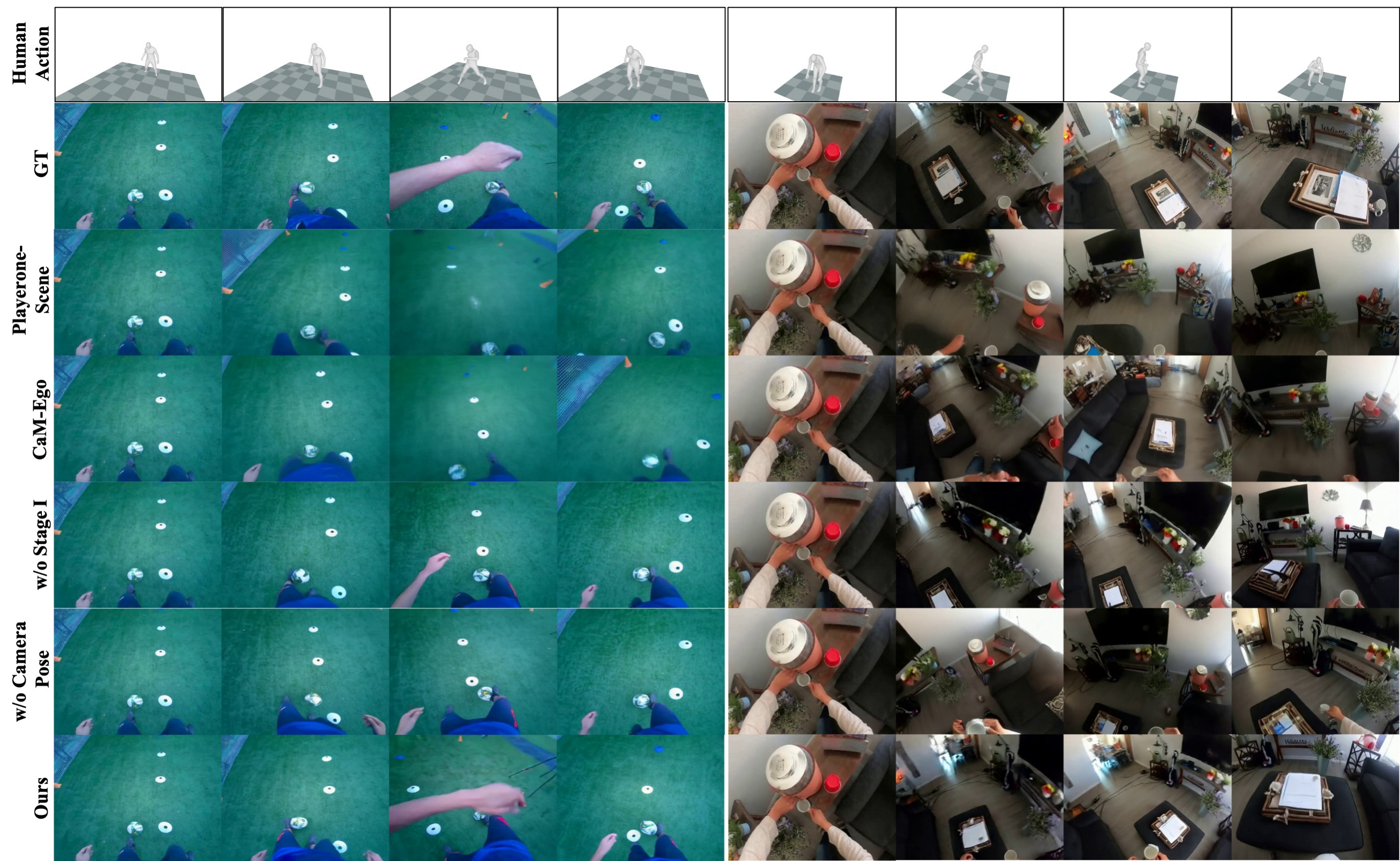}
    \caption{Visualization results of egocentric action control. We show the results compared with baseline methods and our ablation settings.}
    \label{fig:indomain_action}
\end{figure}

\begin{figure}[!t]
    \centering
    \includegraphics[width=\linewidth]{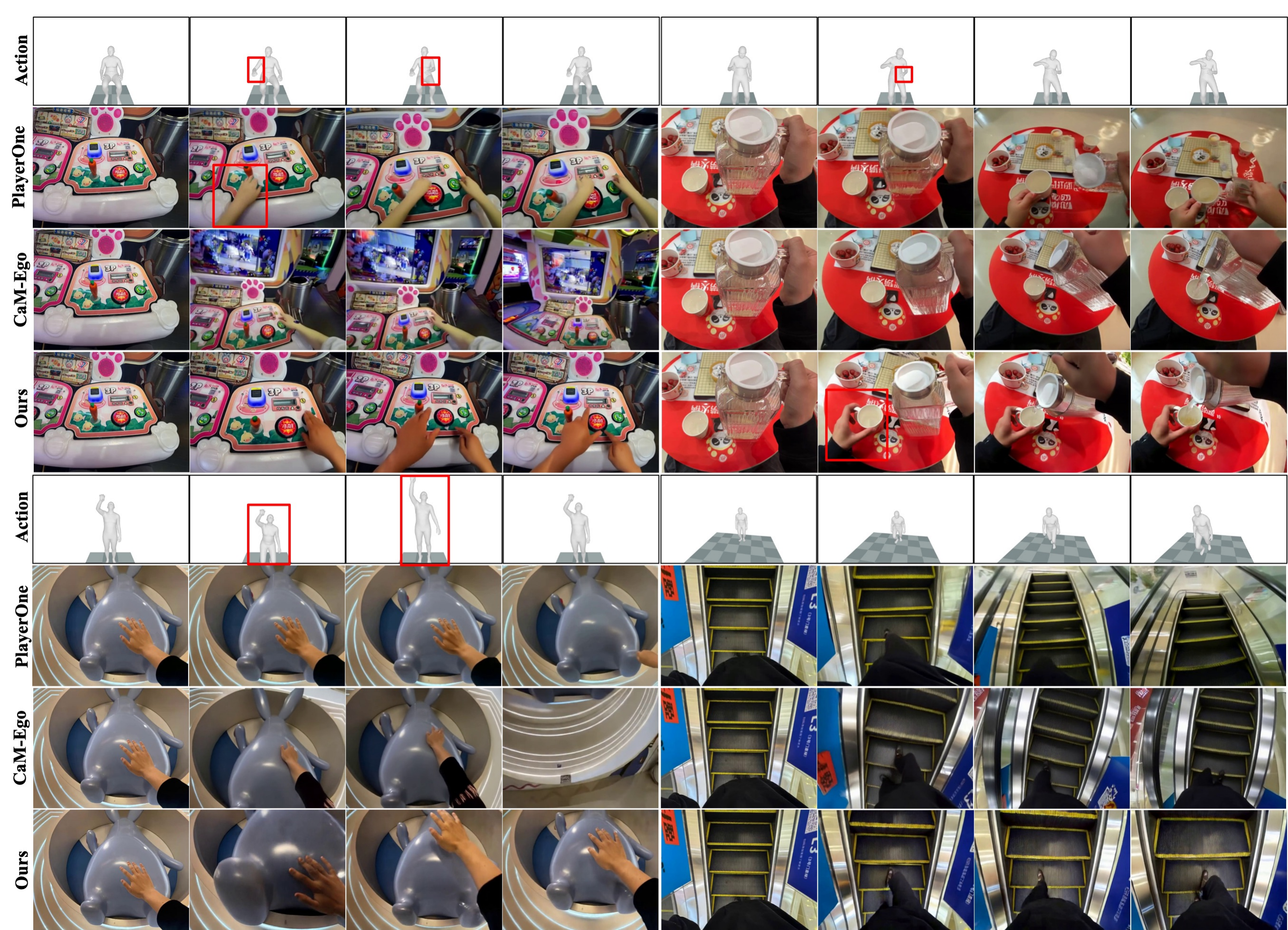}
    \caption{Visualization results of egocentric action control in real-world scenes. Our method generates stable results in response to diverse body motions, such as pouring water, squatting and jumping, and walking up stairs.}
    \label{fig:real_world_action}
\end{figure}

Since most body motions are not visible in egocentric videos, we conduct qualitative comparisons to evaluate the performance of different methods on ego human action control. Figure~\ref{fig:indomain_action} shows the results on the in-domain test set, while Figure~\ref{fig:real_world_action} further compares the results in real-world scenarios. The results demonstrate the superior performance of our projection-based control method. PlayerOne suffers from inaccurate body-motion control, whereas CaM-Ego only supports viewpoint control.

In addition, Figure~\ref{fig:indomain_action} presents a qualitative comparison of the ablation study on the design of egocentric action control. The results show that the absence of motion knowledge from third-person video data, as well as the use of non-projection-based action control, leads to reduced accuracy in body-motion control.

\subsection{Real World Scene}
\begin{figure}[!t]
    \centering
    \includegraphics[width=\linewidth]{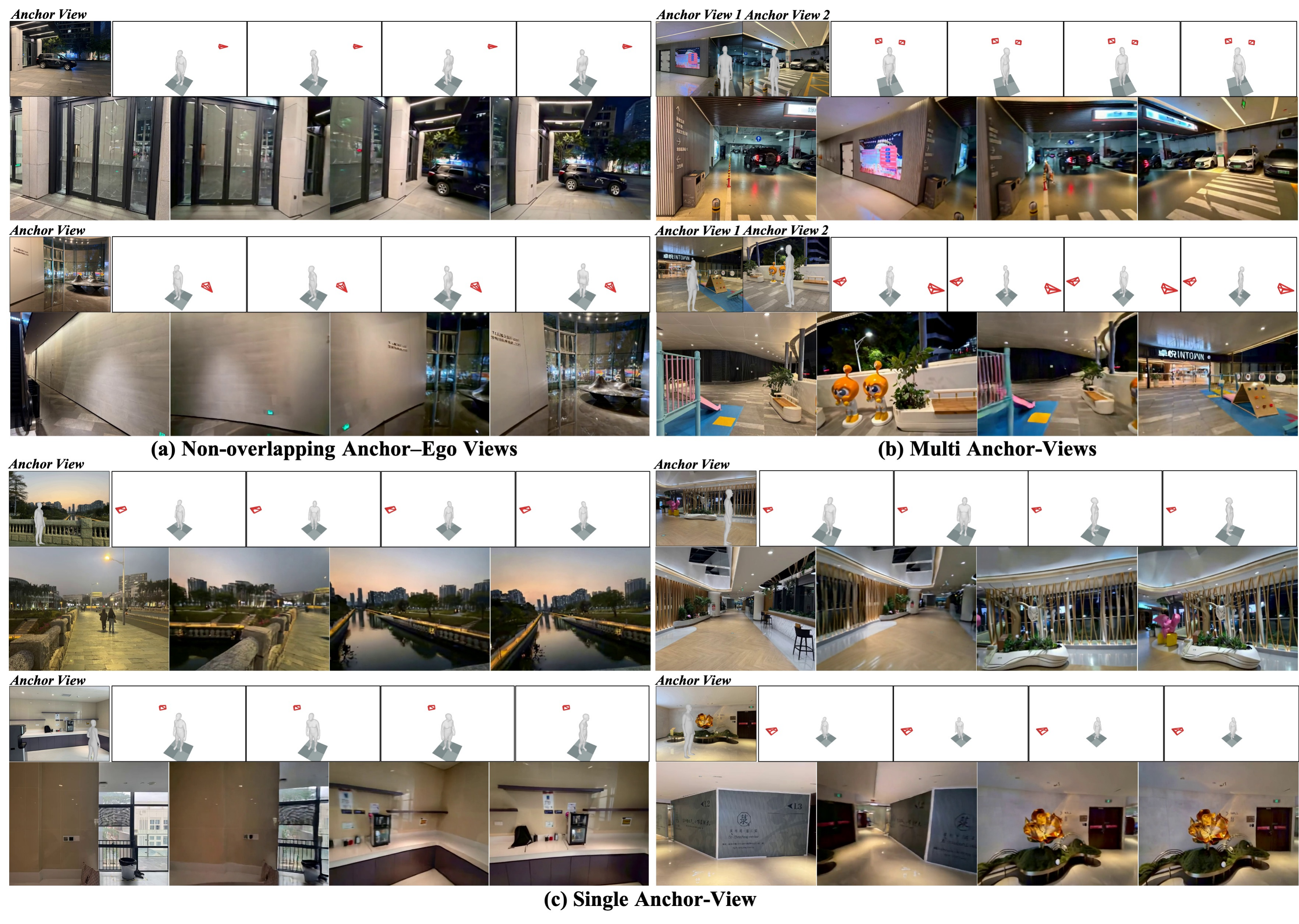}
    \caption{Visualization results in real-world scenes. We show that our method can generate stable results in scenes with non-overlapping viewpoints, as well as in both multi-anchor-view and single-anchor-view settings.}

    \label{fig:real_world_scene}
\end{figure}

To evaluate the generalization ability of our method, we construct test data through real-world capture, as shown in Figure~\ref{fig:real_world_scene}. In addition to the single-anchor-view setting, we construct a multi-anchor-view setting by capturing multiple scene images, where the subject undergoes continuous viewpoint changes that overlap with different anchor views. Furthermore, to verify that our method infers spatial locations from spatial poses rather than relying on overlapping RGB information, we construct test data in which the anchor-view image and the first ego-view frame have no visual overlap by performing coordinate transfer through multiple captures, as illustrated in Figure~\ref{fig:real_world_scene} (a). The results show that our method can still generate correct outputs under this challenging setting, demonstrating its spatial awareness.

\begin{figure}[!t]
    \centering
    \includegraphics[width=\linewidth]{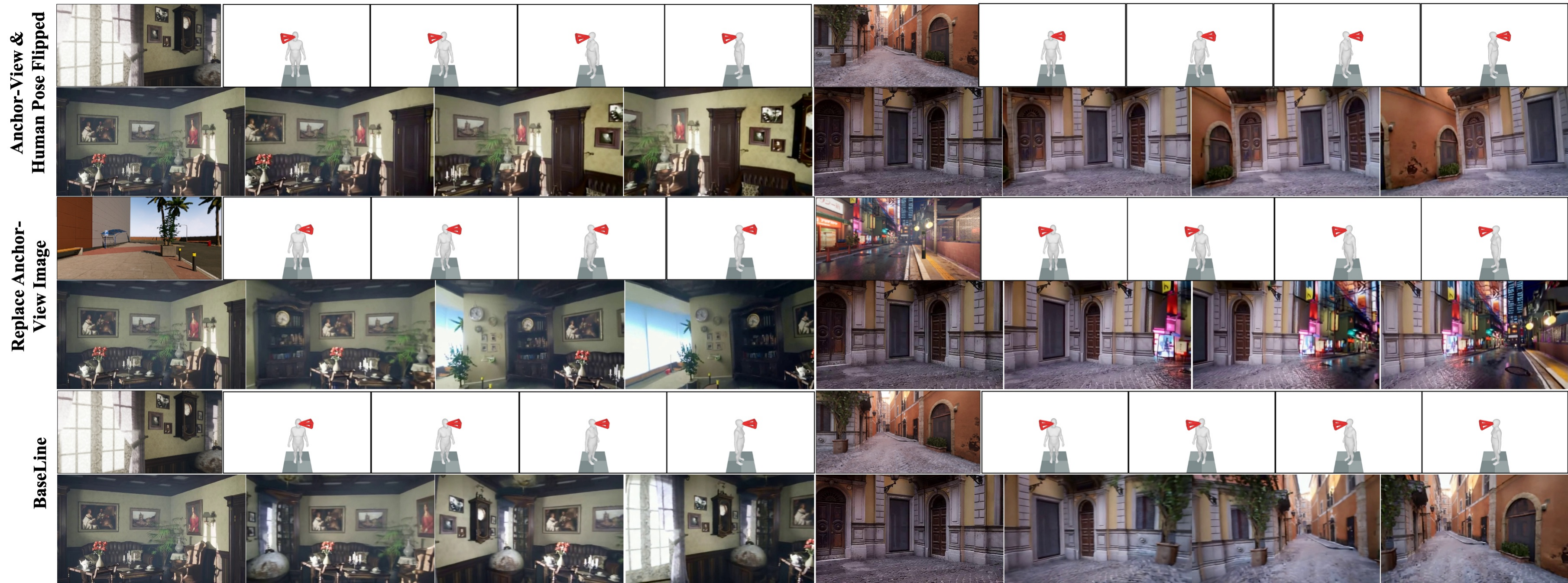}
    \caption{Visualization of scene coherence cases. We replace the anchor view with a style-mismatched image or mirror the world by flipping both the anchor-view and human poses. The results indicate that video generation models internally require a continuous and complete world representation with spatially consistent geometry.}
    \label{fig:scene_coherence}
\end{figure}

\subsection{Scene Coherence.}
As shown in Figure~\ref{fig:scene_coherence}, we consider two challenging settings:
(i) replacing the anchor-view image with another image of a different style, such that the first ego-view frame and the provided anchor scene no longer describe the same underlying world; and
(ii) using the same anchor-view image while simultaneously flipping both the anchor-view pose and the human pose, which mirrors the world space horizontally.
In both settings, the human pose and the anchor-view pose still exhibit apparent view overlap.
However, the generated videos may become inconsistent or visually incoherent.
For setting (i), this is because the model is forced to refer to an anchor scene that is incompatible with the ego-view observation.
For setting (ii), when the world-space geometry becomes inconsistent or physically implausible, the model struggles to generate reasonable results, as can be observed from the wall surface in the first row of Figure~\ref{fig:scene_coherence}.
These results indicate that video generation models internally require a continuous and complete world representation with spatially consistent geometry.

\subsection{Exocentric Action Control}
\begin{figure}[!t]
    \centering
    \includegraphics[width=\linewidth]{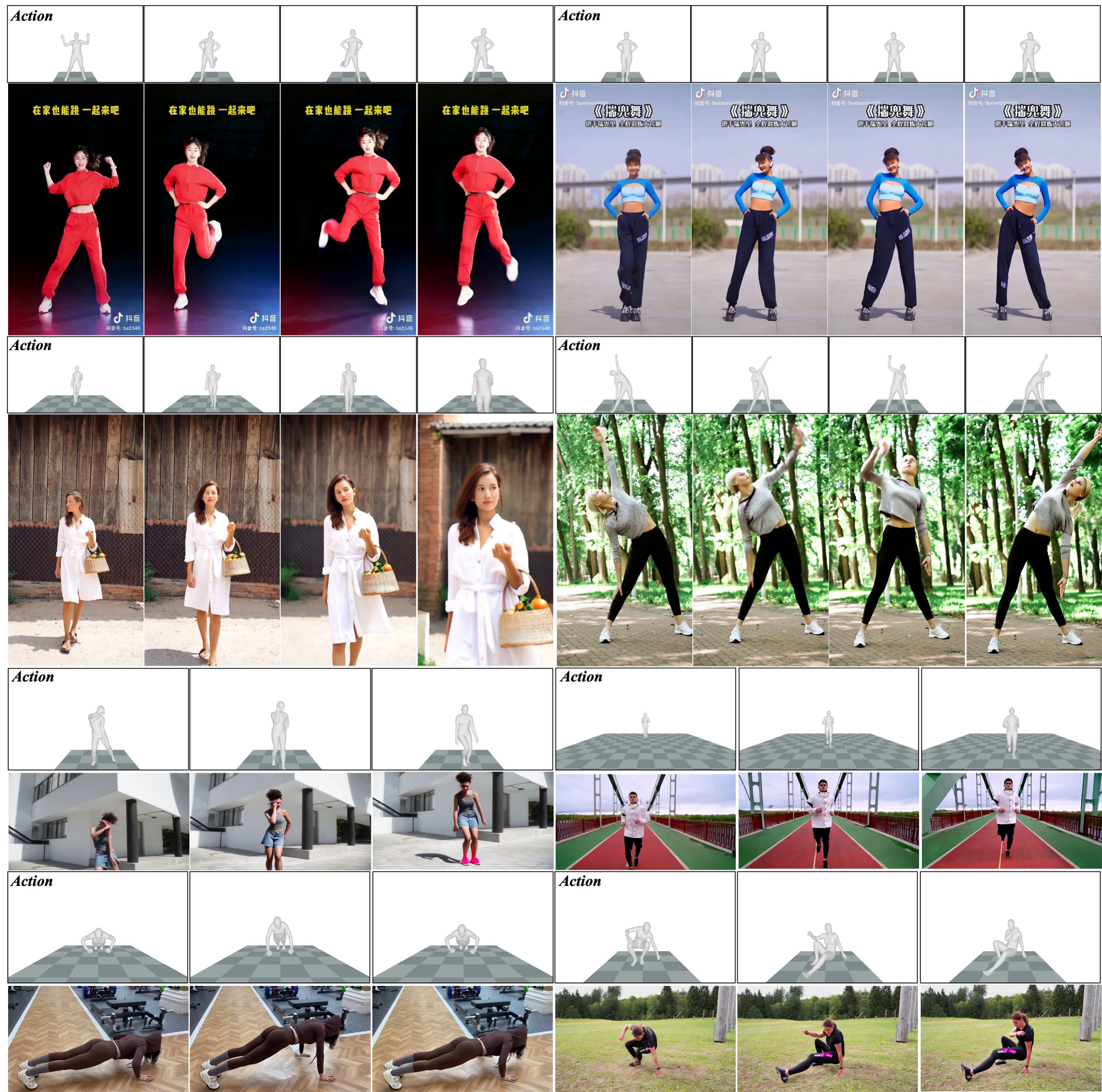}
    \caption{Visualization results of third-person human action control.}

    \label{fig:third_view_action_1}
\end{figure}

We report the ablation results of Stage I third-person human action control in Table~\ref{tab:third_person_action_control_ablation}. We use GVHMR~\cite{shen2024world} to estimate the 3D human poses of the generated videos, and compute MPJPE-related metrics against the ground-truth poses to measure control accuracy. The first row corresponds to using only 3D joint positions to represent joint information, instead of our 6D pose representation. The results show that, due to the lack of orientation information, this design hinders the model from fully understanding the 3D human pose, and may lead to incorrect human orientations in the generated results.

In addition, we explore different pose-condition injection strategies. The results demonstrate that our proposed spatial pose attention achieves the best performance and enables the model to correctly interpret the pose condition. This is because this design explicitly informs the model of the frame-level alignment between video tokens and pose tokens, and drops the pose tokens after attention, since there exists a distribution gap between pose features and VAE latents. Figure~\ref{fig:third_view_action_1} shows visualization results of third-person action control.

\begin{table}[!t]
\centering
\caption{
Quantitative ablation results on \textbf{third-person action control}. 
We report WA-MPJPE and PA-MPJPE, where lower values indicate better performance.
}
\label{tab:third_person_action_control_ablation}
\resizebox{0.55\textwidth}{!}{%
\begin{tabular}{l cc}
\toprule
\textbf{Method} 
& \multicolumn{2}{c}{\textbf{Motion Accuracy}} \\
\cmidrule(lr){2-3}
& WA-MPJPE\(\downarrow\) & PA-MPJPE\(\downarrow\) \\
\midrule

Joint Position Only 
& 90.47 & 38.71 \\

3D Pose Attention 
& 188.17 & 82.37 \\

Cross-Attention Fusion 
& 187.55 & 88.23 \\

In-Context Frame Concat
& 161.67 & 74.64 \\

Ours 
& \textbf{74.57} & \textbf{28.01} \\

\bottomrule
\end{tabular}%
}
\end{table}

\subsection{Additional Quantitative Results}

We show in Table~\ref{tab:scene_image_ablation} how the number of anchor views surrounding a world scene helps improve scene consistency performance. We also present in Table~\ref{tab:vbench_quantitative_comparison} the detailed per-dimension results of the average VBench metrics reported in Table~\ref{tab:quantitative_comparison}.

\begin{table}[!t]
\centering
\caption{
Quantitative ablation results on the number of anchor views. 
}
\label{tab:scene_image_ablation}
\resizebox{0.7\textwidth}{!}{%
\begin{tabular}{c ccccc}
\toprule
\textbf{\# Anchor Views} 
& \multicolumn{5}{c}{\textbf{Scene Consistency}} \\
\cmidrule(lr){2-6} 
& Mat. Pix.(K)\(\uparrow\) & CLIP-V\(\uparrow\) & PSNR\(\uparrow\) & SSIM\(\uparrow\) & LPIPS\(\downarrow\) \\
\midrule

1 &4074.94  &0.8605  &14.9740  &0.5600  &0.5174  \\
2 &4152.91  &0.8645  &15.0294  &0.5585  &0.5178  \\
3 &\textbf{4233.59}  &\textbf{0.8667}  &\textbf{15.1877}  &\textbf{0.5622}  &\textbf{0.5104}  \\

\bottomrule
\end{tabular}%
}
\end{table}

\begin{table*}[!t]
\centering
\caption{
Evaluation metrics cover the fine-grained dimensions of \textbf{VBench}: Subject Consistency (Sub. Cons.), Background Consistency (Bg. Cons.), Temporal Flickering (Temp. Flick.), Motion Smoothness (Mot. Smooth.), Imaging Quality (Img. Qual.), and Aesthetic Quality (Aes. Qual.).
}

\label{tab:vbench_quantitative_comparison}
\resizebox{\textwidth}{!}{%
\begin{tabular}{l cccccc}
\toprule
\textbf{Method} 
& \multicolumn{6}{c}{\textbf{VBench Dimensions}} \\
\cmidrule(lr){2-7} 
& Sub. Cons.\(\uparrow\) & Bg. Cons.\(\uparrow\) & Temp. Flick.\(\uparrow\) & Mot. Smooth.\(\uparrow\) & Img. Qual.\(\uparrow\) & Aes. Qual.\(\uparrow\) \\
\midrule

\multicolumn{7}{l}{\textbf{Ego Static Scene}} \\ 
\midrule
PlayerOne             &0.7956  &0.8964  &0.9474  &0.9821  &0.3945  &0.3800  \\
PlayerOne-Scene       &0.8071  &0.8974  &0.9498  &0.9820  &0.3940  &0.3803  \\
CaM-UE                &0.7694  &0.8899  &0.9357  &0.9811  &0.4099  &0.3903  \\
CaM-Ego               &0.8142  &0.9040  &\textbf{0.9533}  &\textbf{0.9851}  &\textbf{0.4172}  &0.4155  \\
\textbf{Ours}         &\textbf{0.8167}  &\textbf{0.9041}  &0.9523  &0.9832  &0.4140  &\textbf{0.4171}  \\
\midrule

\multicolumn{7}{l}{\textbf{UE CineScene}} \\ 
\midrule
PlayerOne             &0.7920  &0.8600  &0.9361  &0.9818  &0.4309  &0.4125  \\
PlayerOne-Scene       &0.8147  &0.8699  &0.9444  &0.9856  &0.4026  &0.4052  \\
CaM-UE                &0.8004  &0.8961  &0.9289  &0.9903  &0.4214  &0.4631  \\
CaM-Ego               &0.8496  &\textbf{0.9035}  &\textbf{0.9426}  &\textbf{0.9911}  &0.4566  &\textbf{0.4789}  \\
\textbf{Ours}         &\textbf{0.8522}  &0.8986  &0.9382  &\textbf{0.9911}  &\textbf{0.4571}  &0.4781  \\
\midrule

\multicolumn{7}{l}{\textbf{Ego Dynamic Scene}} \\ 
\midrule
PlayerOne-Scene       &0.8743  &0.9140  &0.9649  &0.9889  &0.4015  &0.3941  \\
CaM-UE                &0.8824  &0.9230  &0.9586  &\textbf{0.9921}  &\textbf{0.4508}  &0.4156  \\
CaM-Ego               &0.8751  &0.9266  &0.9669  &0.9913  &0.4388  &0.4204  \\
\textbf{Ours}         &\textbf{0.8937}  &\textbf{0.9295}  &\textbf{0.9689}  &0.9901  &0.4371  &\textbf{0.4272}  \\

\bottomrule
\end{tabular}%
}
\end{table*}

\section{Failure Cases}\label{app:d}

\begin{figure}[!t]
    \centering
    \includegraphics[width=\linewidth]{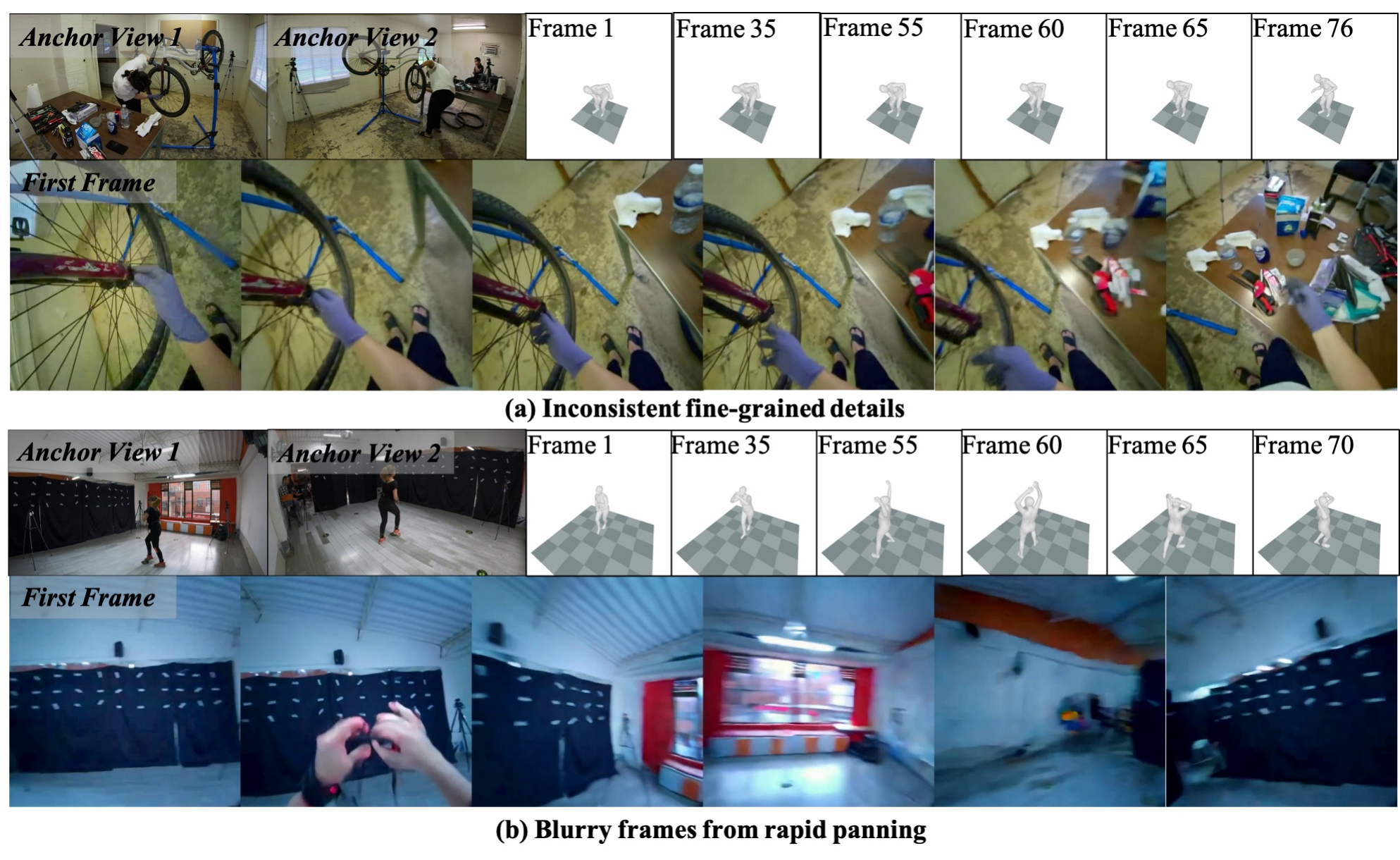}
    \caption{Failure cases. (a) Due to the limited capability of the base model, our method may struggle to preserve highly fine-grained texture details in scenes with complex local structures, leading to inconsistent scene details. (b) Since egocentric videos often involve rapid viewpoint changes, the training data contains blurry frames, which may result in blurry generation artifacts.}
    \label{fig:failure_case}
\end{figure}

\noindent \textbf{Inconsistent Scene Details.}
We observe that when local regions of a scene contain complex structures and rich texture details, our method may produce results with inconsistent fine-grained details, as shown in Figure~\ref{fig:failure_case} (a). We believe that this limitation is largely constrained by the capability of the base model. Specifically, the VAE of Wan TI2V 2.2 5B used in our experiments has a spatial downsampling factor of 16, leading to a relatively high compression ratio in the latent spatial dimensions and thus the loss of fine-detail information. In the future, adopting more powerful base models is expected to alleviate this issue.

\noindent \textbf{Blurry Results.}
Our first-person training data contains a large number of videos with rapid viewpoint changes, which often leads to motion blur in the frames. Consequently, the generated results may also exhibit similar blurring artifacts, as shown in Figure~\ref{fig:failure_case} (b). In addition, due to the limitations of the base model and the fast motion commonly present in first-person data, the generated hands may suffer from degraded visual quality.

\begin{table*}[!ht]
\caption{Instruction Template for Evolution Prompt}
\label{tab:fpv_character_description_template}

\centering
\footnotesize

\begin{minipage}{1.0\columnwidth}
\begin{tcolorbox}
\begin{tabular}{p{0.97\columnwidth}}

\VarSty{{\bf Role}} \\
You are an expert video analyst specializing in First-Person Perspective (FPV) footage.\\

\VarSty{{\bf Objective}} \\
Your primary task is to detect and describe the \textbf{external character(s)} visible in the video.
Do not describe the first-person observer, the camera wearer, or their body parts.\\

\VarSty{{\bf Detection Phase}} \\
Scan the video for distinct people other than the camera wearer.\\
\quad If no external person is present, or if people are too obscured to be identifiable, strictly output: \texttt{False}.\\
\quad If distinct external people are visible, proceed to the description phase.\\

\VarSty{{\bf Description Strategy}} \\
Select the subject to describe according to the following rules:\\
1. \textbf{Main subject:} If one person is the clear focal point, such as interacting with the camera wearer, being closest to the observer, or performing a distinctive action, describe only this main person and ignore background extras.\\
2. \textbf{Collective group:} If multiple people are present without a clear protagonist, describe them as a single collective unit, such as ``a group of students'' or ``a crowd of pedestrians''.\\
3. \textbf{Constraint:} Do not enumerate individuals as ``Character 1'', ``Character 2'', or ``Character 3''. Describe either the main subject or the collective group state.\\

\VarSty{{\bf Description Aspects}} \\
Provide a concise and objective description covering the following four aspects:\\
\quad \textbf{a) Location \& Position:} Describe the subject's spatial location in the physical 3D environment. Relate the position to physical objects or spaces, such as ``sitting on the sofa'', ``leaning against the wall'', ``standing in the doorway'', or ``walking down the hallway''. Do not describe positions relative to the video frame, such as ``in the center of the screen'', ``on the left side'', or ``bottom right''.\\
\quad \textbf{b) Appearance:} Describe visible physical traits and clothing. For groups, describe the shared appearance, such as suits or casual clothes.\\
\quad \textbf{c) Dynamic Actions:} Describe the subject's movement or activity, focusing on the flow of motion.\\
\quad \textbf{d) Interaction with ``I'':} Briefly state the interaction type if it exists, such as ``talking to me'', ``handing an object'', or ``blocking my path''. If no direct interaction is observed, state ``None''.\\

\VarSty{{\bf Constraints}} \\
Describe only visually confirmed content and avoid hallucinations. Keep the description concise and do not list multiple individuals separately.\\

\VarSty{\textbf{Output example:}}\\
\textbf{a) Location \& Position:} The person is standing in the living room, positioned between a coffee table and a TV console.\\
\textbf{b) Appearance:} The person is wearing a black short-sleeved T-shirt with a white graphic, gray knee-length shorts, flip-flops, and a backward black cap.\\
\textbf{c) Dynamic Actions:} The person is vacuuming the floor, pushing the upright vacuum cleaner forward and then bending down.\\
\textbf{d) Interaction with ``I'':} None. The person is engaged in cleaning and does not appear to interact with the camera wearer.\\

\end{tabular}
\end{tcolorbox}
\end{minipage}
\end{table*}

\end{document}